\documentclass{article} 

\PassOptionsToPackage{table,dvipsnames}{xcolor}
\usepackage{preprint,times}

\usepackage{array}
\usepackage{amsmath}
\usepackage{amsfonts}
\usepackage{amssymb}
\usepackage{nicefrac}
\usepackage{nicematrix}
\usepackage{hhline}
\newcolumntype{R}{>{$}r<{$}}
\newcolumntype{C}{>{$}c<{$}}

\usepackage[
    detect-all,
    separate-uncertainty=true
]{siunitx}

\usepackage{amsmath,amsfonts,bm}

\def\eqref#1{equation~\ref{#1}}

\def\1{\bm{1}}

\def\vx{{\bm{x}}}

\def\mI{{\bm{I}}}

\DeclareMathAlphabet{\mathsfit}{\encodingdefault}{\sfdefault}{m}{sl}
\SetMathAlphabet{\mathsfit}{bold}{\encodingdefault}{\sfdefault}{bx}{n}

\usepackage{graphicx}
\usepackage{booktabs}
\usepackage{makecell}
\usepackage{multirow}
\usepackage{tabularx}
\usepackage{subcaption}
\usepackage{caption}
\usepackage[table,dvipsnames]{xcolor}

\usepackage{microtype}
\usepackage{cuted}
\usepackage{enumitem}
\usepackage{pifont}

\usepackage[most]{tcolorbox}

\usepackage{titletoc}
\newcommand\DoToC{%
  \startcontents
  \printcontents{}{1}{\noindent \textbf{\Large{Table of Contents in Appendix}}\vskip3pt\vskip5pt}
  \vskip3pt\vskip5pt
}
\usepackage{placeins}

\usepackage{url}

\usepackage[
    colorlinks=true,
    citecolor=BlueViolet,
    linkcolor=purple,
    urlcolor=BlueViolet
]{hyperref}

\newcommand{\mypara}[1]{%
    \noindent\textbf{#1.}\xspace
}

\usepackage{xspace}

\title{Mind the Tail:\\ Data Selection Through Sparse Concepts}

\author{%
  Ziheng Zhang$^{1,2\dagger}$\; Neha Hulkund$^{3}$\; Colin Lee$^{2}$\; Zheda Mai$^{2}$\; Hui Shen$^{4}$\\
  \textbf{Ying Ding$^{1}$\; Tanya Berger-Wolf$^{2}$\; Sara Beery$^{3}$\; Wei-Lun Chao$^{5}$\; Jianyang Gu$^{2,3\dagger}$}\\[3pt]
  {\small $^{1}$University of Texas at Austin\quad $^{2}$The Ohio State University\quad $^{3}$Massachusetts Institute of Technology}\\
  {\small $^{4}$University of Michigan\quad $^{5}$Boston University}\\
  {\small $^{\dagger}$\texttt{zihengzhang@utexas.edu}, \texttt{gu.1220@osu.edu}}
}

\begin{document}

\maketitle

\begin{abstract}
Data selection reallocates training support across visual concepts, including objects, attributes, color, texture, and backgrounds. Different allocation strategies may produce varied downstream model behavior, especially for tail concepts supported by only a few samples. Discarding them can hurt test-time robustness, while overemphasizing samples lying on category boundaries may introduce unstable training signals. This work explicitly investigates this relationship across fine-grained, spurious-correlation, and general image classification, as well as multimodal pretraining. Specifically, we use sparse autoencoders (SAEs) to extract concept activations from individual images. During selection, we accumulate the dataset-level concept support for each added sample. Applying diminishing returns to this distribution down-weights well-represented concepts and shifts the focus to the distribution tail. We then incorporate local support to find samples that reliably strengthen under-supported concepts. Empirical results show consistent improvements over strong data selection baselines across all benchmarks. We also reveal the connection between concepts our method enhances and the induced behavioral changes, including improved robustness against out-of-distribution and spurious contexts.
This work establishes concept support, beyond individual sample importance, as a useful principle for data-efficient and robust learning\footnote{We attach source code, selected subsets, and models at \href{https://imageomics.github.io/mind-the-tail/}{imageomics.github.io/mind-the-tail}.}. 
\end{abstract}

\section{Introduction}

Data selection reduces training costs by restricting optimization to a subset of available samples~\citep{guo2022deepcore,toneva2018forgetting,mirzasoleiman2020coresets}. Retaining samples that produce strong training signals or capture local statistics likely preserves model performance under a reduced training budget~\citep{sener2018active,paul2021gradn,zheng2023coverage}. Yet data selection also plays a more proactive role: \textit{data shapes model behavior}. Machine learning models continuously pick up bias from training data, which can include representative traits for recognizing a bird species but can also include observation-induced background elements, such as bird feeders, creating potential spurious correlations~\citep{torralba2011unbiased,dharmasiri2025impact}. 

\begin{figure}[t]
    \centering
    \includegraphics[width=\linewidth]{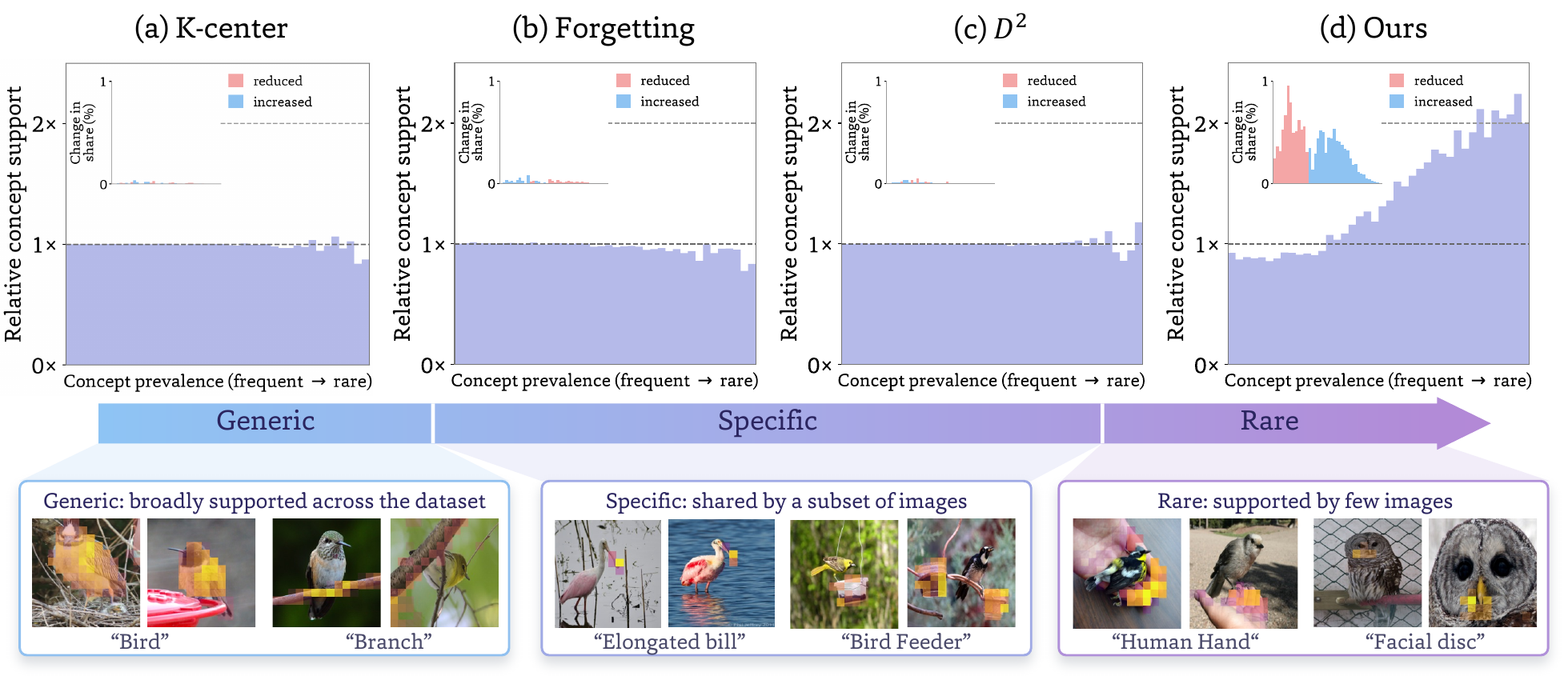}
    \vspace{-16pt}
\caption{\textbf{Concept-support distribution produced by different selectors.}
Data selection reallocates the concept distribution from the original dataset. The top bar compares three previous methods and our approach on the relative change in concept support in a $10$\% subset. Our method substantially amplifies tail concepts (insets show the absolute change, where red is suppression and blue is amplification). In contrast, K-Center, Forgetting, and D$^2$ Pruning largely preserve the original concept distribution.
Bottom: representative SAE concepts across the same prevalence spectrum, from broadly supported \textbf{generic} concepts, through \textbf{specific} patterns shared by a subset of images, to \textbf{rare} concepts supported by only a few images. Highlighted patches indicate high-activation regions.}
    \label{fig:teaser}
    \vspace{-8pt}
\end{figure}

Such a close relationship between the prevalence of visual evidence and downstream behavior motivates us to rethink data selection through the lens of \textit{concepts}. Each image carries various concepts, including objects, attributes, textures, and backgrounds~\citep{choi2025conceptscope}, and their support is inherently uneven across a dataset~\citep{zhu2014longtail,liu2019oltr}. As shown in Figure~\ref{fig:teaser}, generic concepts, such as bird and branch, are broadly supported by many samples; toward the long tail, we observe appearance and background concepts associated with fewer categories or incidental events. Consequently, selecting subsets implicitly reallocates training support across all of the concepts the samples carry.
More broadly, we argue that data selection \textit{need not preserve the source distribution}; it should allocate training support to evidence most useful for downstream learning. While this intuition has been recognized in targeted data selection~\citep{xia2024less,engstrom2024dsdm,xie2023dsir}, long-tailed selection~\citep{vo2024automatic,xu2024metaclip}, and other specialized settings, we still lack a principled way to determine the concept allocation across different learning scenarios.

We study this question using concepts derived by Sparse autoencoders (SAEs), which decompose pretrained features into sparse coordinates associated with recurring visual patterns~\citep{makhzani2014ksparse,gao2025scaling,choi2025conceptscope,stevens2025interpretable}. The sparse representation characterizes a selected subset by its induced concept support.
Unlike previous data-selection literature based on SAEs~\citep{yang2025diversity,zhou2026dynamic,das2026matched}, we seek a direct relationship between the allocation of training support across concepts and the resulting downstream model behavior.
More specifically, different source distributions, including fine-grained, spurious-correlation, and general datasets, can have distinct demands on the concept tail: some settings may favor repeatedly supported discriminative evidence, while others require greater support for less frequent concepts. 

To this end, we formulate a concept-based data-selection objective to make concept allocation explicit.
Specifically, we accumulate a concept distribution as samples are selected. Imposing diminishing returns on this distribution gradually shifts selection pressure from saturated concepts toward the under-supported tail. After identifying valuable concepts that fill the distribution gap, we use local support to ensure that the selected samples, as well as their neighbors, contribute to those concepts. Therefore, the selection process favors recurring underrepresented concepts over isolated occurrences. 
Together, these components allow us to vary the balance between local support and tail concepts and study the downstream behaviors induced by different allocation strategies.

As shown in Figure~\ref{fig:teaser}, our method significantly alters the concept distribution in the selected subset, particularly amplifying tail concepts.
Empirically, our selected subsets consistently outperform strong data-selection baselines across classification and multimodal pretraining tasks. More importantly, varying concept allocation reveals systematic changes in downstream behavior. Concept support enhanced by the proposed strategy improves test accuracy on data with those concepts.
Even for concepts not directly tied to specific categories, increasing exposure during training reduces the risk of test samples falling out of distribution. 
Broader concept coverage can improve robustness to spurious context, while aggressively prioritizing tail concepts can hurt performance when recurring discriminative evidence matters more. These findings establish concept allocation, beyond individual sample importance, as a useful interface for studying and designing data selection.

\section{Related Work}

\mypara{Dataset selection}
Data selection, or data pruning, seeks to retain a compact training subset
while preserving downstream performance~\citep{guo2022deepcore}.
Existing methods characterize useful data from several perspectives.
Geometry-based approaches emphasize representativeness and diversity through
clustering, K-Center, or submodular coverage
~\citep{sener2018active,wei2015submodularity}, while training-dynamics and
optimization-based methods use forgetting events, early-training scores,
gradients, or validation objectives
~\citep{toneva2018forgetting,paul2021gradn,mirzasoleiman2020coresets,
killamsetty2021gradmatch,killamsetty2021glister}.
Self-supervised scores further enable pruning without supervised scoring models
~\citep{sorscher2022beyond}.
Recent methods increasingly combine sample utility with coverage and
redundancy, including Moderate Coreset, CCS, D$^2$-Pruning, and InfoMax
~\citep{xia2023moderate,zheng2023coverage,maharana2024d2,tan2025infomax},
while data-attribution methods estimate the contribution of individual
training examples~\citep{koh2017influence,ghorbani2019data}.
Pretrained representations have further enabled fine-grained and multimodal
data selection, including RAM-APL and large-scale image--text filtering
~\citep{wan2025foundation,gadre2023datacomp,fang2024data,
mahmoud2024sieve,nicolicioiu2026data}. Concept bottlenecks have also been used to estimate sample difficulty for
coreset selection~\citep{mehra2025model}. Our method performs data selection through sparse concepts across fine-grained, spurious-correlation, and general settings. Importantly, we examine the relationship between concept distributions produced by different selection strategies and downstream model behavior across dataset types.

\mypara{Concept representations and sparse autoencoders}
Concept-based interpretability aims to describe model behavior using
human-understandable factors, as explored by methods such as Network Dissection,
TCAV, ACE, and concept bottleneck models
~\citep{bau2017network,kim2018tcav,ghorbani2019ace,koh2020concept}.
Sparse autoencoders (SAEs) provide a complementary approach by decomposing
dense activations into sparse, often interpretable features
~\citep{cunningham2023sparse,bricken2023monosemanticity,gao2025scaling}.
Recent work extends these representations to vision and vision--language
models, where SAE features capture localized and hierarchical visual patterns
~\citep{stevens2025interpretable,pach2025sparse,bussmann2025matryoshka,
zaigrajew2025interpreting}.
Concept representations have also begun to support data selection, including
SDM for target-distribution matching, alongside work on concept-based difficulty
estimation, diversity-driven selection, and dynamic concept exposure
~\citep{das2026matched,mehra2025model,yang2025diversity,zhou2026dynamic}.
Our work directly optimizes concept support together with local support,
yielding an interpretable selection criterion for classification and
multimodal pretraining.

\section{Methodology}\label{sec:methodology}
Let $\mathcal{D}=\{I_i\}_{i=1}^{n}$ be a set of training images and
$B \ll n$ a budget. Data selection constructs a subset $\mathcal{S} \subseteq \mathcal{D}$ with
$|\mathcal{S}| = B$ that preserves the information in the original dataset needed for model training. 
Each image consists of various concepts, such as objects, attributes, colors, and textures. Therefore, we view the data selection process as reallocating support for these concepts under a limited budget. Specifically, we apply an SAE with $C$ dimensions to the DINOv3~\citep{simeoni2025dinov3} patch features of each image and max-pool the resulting patch activations, so that each image $\mI_i$ is represented by a sparse vector $\vx_i\in\mathbb{R}^{C}$, where $\vx_i[c]$ is the corresponding activation for concept $c$, \textit{i.e.}, the $c$-th SAE dimension.

SAE-extracted concepts usually demonstrate a long-tailed distribution. A small number of concepts receive support from many images, while other concepts occur in only a limited portion of the dataset. These tail concepts can capture fine-grained attributes and uncommon objects or context variations. Although the latter may not be directly tied to a category, removing the few samples that contain them can narrow the training distribution's support, making similar test cases more likely to appear out of distribution. However, their limited support also makes these concepts particularly vulnerable to data reduction, motivating us to explicitly track the accumulated concept support in the selected subset.

\begin{figure}[t]
    \centering
    \includegraphics[width=\linewidth]{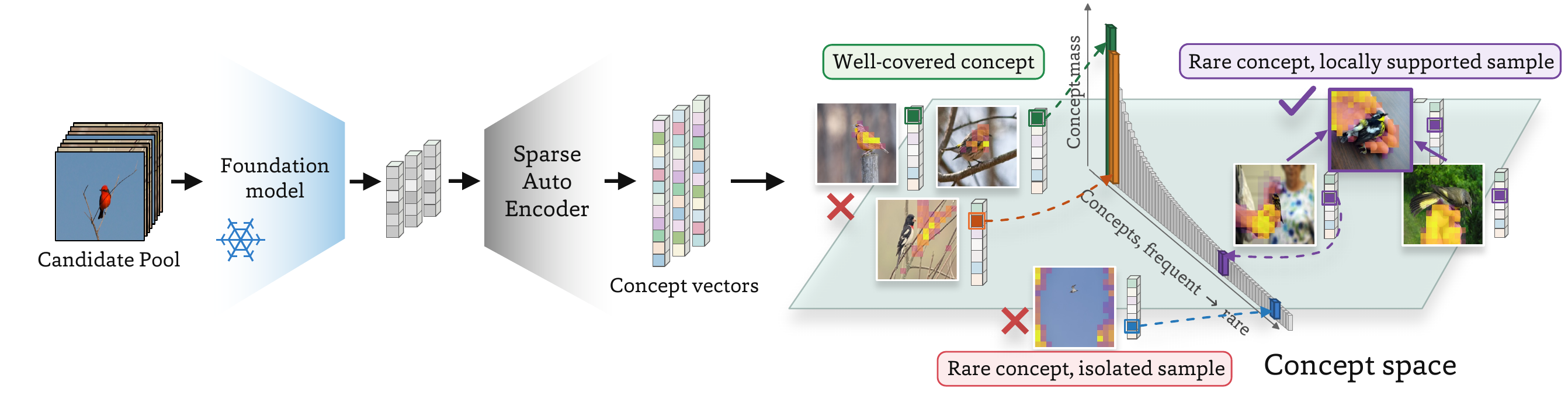}
    \vspace{-14pt}
\caption{
\textbf{Overview of concept-based subset selection.}
We use SAEs to decompose foundation-model patch features into sparse concept vectors and track the accumulated support of each concept in the selected subset.
Concept-level diminishing returns reduce the marginal value of already well-covered evidence, steering the selection toward tail concepts. We then use local support to locate samples that contribute to filling the distributional gap together with their neighbors.
The resulting objective favors candidates representing under-supported evidence
with local support over redundant head concepts or isolated rare regions.
}
    \label{fig:pipeline}
    \vspace{-8pt}
\end{figure}

At the same time, the attention to tail concepts does not imply blindly equalizing the concept distribution. Different source distributions place varied demand on concept allocation under a limited budget. Expanding concept coverage too aggressively toward the tail may introduce chaotic training signals from isolated samples. We therefore formulate data selection as a progression from \textit{concept value} to \textit{sample value}: it should identify concepts that are insufficiently represented as selection proceeds, while also determining which samples provide reliable support for those concepts in the source distribution. Our method follows this progression, measuring \textbf{concept coverage} to suggest \textit{what evidence} is needed and \textbf{local support} to decide \textit{which samples} meaningfully fill that gap. Figure~\ref{fig:pipeline} illustrates the proposed selection pipeline.

\subsection{Concept Coverage as a Probe for the Distribution Tail}
\label{sec:concept_objective}

We first apply diminishing returns to accumulated concept support and score a subset by
\begin{equation}
    F(\mathcal{S})
    =
    \sum_{c=1}^{C}
    \log\!\left(
        1+
        \sum_{\mI_i\in \mathcal{S}}\vx_i[c]
    \right).
    \label{eq:objective}
\end{equation}

The logarithm reduces the marginal gain of repeatedly adding evidence for concepts that are already well represented. Consequently, under-supported concepts become more competitive as selection proceeds, without an explicit definition of the tail or a target concept frequency. 
\S\ref{sec:obj_design} compares alternative concave utilities.

\subsection{Local Support to Expose Valuable Samples}
Concept coverage determines \textit{where} additional support is needed, but does not by itself indicate \textit{whether} a sample carrying an under-supported concept reflects meaningful structure in the source distribution. Therefore, we augment each candidate with support from its local neighborhood. For each image $\mI_t$, let $\mathcal{N}_k(t)$
denote its $k$ nearest neighbors in SAE space. We accordingly update the concept contribution in Eq.~\ref{eq:objective} for candidate $\mI_i$ to
\begin{equation}
    \tilde{\vx}_i[c]
    =
    \vx_i[c]
    +
    \beta
    \sum_{t:\,\mI_i\in\mathcal{N}_k(t)}
    a_{ti}\vx_t[c],
    \label{eq:local}
\end{equation}
where $\beta$ controls the strength of local support and $a_{ti}$ is the cosine similarity between $\mI_t$ and $\mI_i$, normalized across the neighboring set $\mathcal{N}_k(t)$.
The first term captures concepts directly present in $I_i$, and the second term assigns complementary credit recurring among nearby samples. As a result, repeatedly supported concepts gradually receive diminishing marginal gains. Yet rare concepts do not automatically dominate selection because the corresponding samples may receive limited local support when isolated.
We report ablations of $\beta$ and the row normalization of $a_{ti}$ in \S\ref{sec:ablation}, with additional neighborhood-size and objective-design studies in \S\ref{sec:obj_design}.

\mypara{Optimization}
The objective also admits efficient optimization.
Since $\log(1+\cdot)$ is monotone and concave and $\tilde{x}_i \ge 0$, $F$ is monotone submodular, greedy selection attains a $(1-1/e)$ approximation to the optimal subset~\citep{nemhauser1978analysis}.
We run lazy greedy~\citep{minoux1978accelerated} from $\mathcal{S}=\emptyset$, avoiding recomputing the marginal gain of every candidate at each iteration. Because both the original SAE activations and their effective concept contributions are sparse, each marginal-gain computation only involves the nonzero dimensions of $\tilde{\vx}_i$. The resulting procedure preserves the structure of the concept-based objective while remaining practical for large-scale subset selection.

\section{Experiments}
\mypara{Tasks, models, and datasets}
We evaluate our method on image classification and multimodal pretraining. For image
classification, we use NABirds~\citep{vanhorn2015building} and Stanford
Cars~\citep{krause2013collecting} for fine-grained recognition,
Waterbirds~\citep{sagawa2020distributionally} and
Spawrious~\citep{lynch2023spawrious} for robustness to spurious correlations.
We train ResNet-50~\citep{he2016deep} on the selected subsets. For multimodal
pretraining, we select subsets from CC12M~\citep{changpinyo2021conceptual} and train
CLIP ViT-B/16~\citep{clip2021}. We evaluate ImageNet-1K zero-shot and linear-probe classification,
together with image-to-text and text-to-image retrieval on
Flickr30K~\citep{plummer2015flickr30k}. All methods use the same selection budget
and downstream training configuration.

\mypara{Baselines}
In the classification task, we compare with
\textbf{1) Random}, uniform subset sampling;
\textbf{2) K-Center}~\citep{sener2018active}, geometry-based feature-space coverage;
\textbf{3) Forgetting}~\citep{toneva2018forgetting}, based on forgetting events during training;
\textbf{4) EL2N}~\citep{paul2021gradn}, based on prediction-error norms;
\textbf{5) Moderate}~\citep{xia2023moderate}, which selects samples of moderate difficulty;
\textbf{6) CCS}~\citep{zheng2023coverage}, coverage-aware sampling over difficulty scores;
\textbf{7) D$^2$-Pruning}~\citep{maharana2024d2}, which combines difficulty and diversity;
\textbf{8) InfoMax}~\citep{tan2025infomax}, which jointly considers informativeness and redundancy; and
\textbf{9) RAM-APL}~\citep{wan2025foundation}, foundation-model-based selection for
fine-grained recognition. \textbf{10) SDM}~\citep{das2026matched}, submodular distribution matching over SAE concepts. We follow the original supervision and implementation protocols of each method. For multimodal pretraining, we use the CLIP score~\citep{hessel-etal-2021-clipscore} as a strong baseline for CC12M.
Based on the CLIP scores, we instantiate \textbf{Moderate}, \textbf{CCS},
\textbf{D$^2$-Pruning}, and \textbf{InfoMax}. We additionally compare with
\textbf{Random} and \textbf{K-Center} as unsupervised baselines.
\begin{table}[tp]

\caption{
\textbf{Results on image classification datasets.}
Comparison of subset selection performance with existing coreset selection methods using ResNet-50.
We report classification accuracy for NABirds and Stanford Cars, and worst-group accuracy (WGA) for Waterbirds and Spawrious. All results are averaged over three  runs.
\textbf{Bold} and \underline{underlined} indicate the best and second-best results, respectively.
}

\label{tab:cls_results}

\vspace{-0.2cm}

\small
\setlength{\tabcolsep}{2.0pt}
\centering

\resizebox{\linewidth}{!}{
\begin{tabular}{l|cccccc|cccccc|cccccc|cccccc}

\toprule[1pt]

\textbf{Methods}
& \multicolumn{6}{c|}{\textbf{NABirds}}
& \multicolumn{6}{c|}{\textbf{Stanford Cars}}
& \multicolumn{6}{c|}{\textbf{Waterbirds}}
& \multicolumn{6}{c}{\textbf{Spawrious}} \\

\cmidrule(lr){2-7}
\cmidrule(lr){8-13}
\cmidrule(lr){14-19}
\cmidrule(lr){20-25}

\textbf{Selection Rate}
& \textbf{10\%} & \textbf{20\%} & \textbf{30\%} & \textbf{50\%} & \textbf{70\%} & \textbf{100\%}
& \textbf{10\%} & \textbf{20\%} & \textbf{30\%} & \textbf{50\%} & \textbf{70\%} & \textbf{100\%}
& \textbf{10\%} & \textbf{20\%} & \textbf{30\%} & \textbf{50\%} & \textbf{70\%} & \textbf{100\%}
& \textbf{10\%} & \textbf{20\%} & \textbf{30\%} & \textbf{50\%} & \textbf{70\%} & \textbf{100\%} \\

\midrule[1pt]

\rowcolor{gray!15}
\textbf{Random}
& $35.6$ & $55.3$ & $64.9$ & $73.3$ & $77.1$ & $80.3$
& $21.9$ & $45.2$ & $63.1$ & $\underline{78.0}$ & $83.7$ & $87.5$
& $52.3$ & $66.2$ & $67.5$ & $68.8$ & $\underline{79.1}$ & $80.1$
& $63.8$ & $67.4$ & $72.6$ & $68.8$ & $69.5$ & $76.3$ \\

\midrule[0.1pt]

\textbf{K-Center}
& $31.6$ & $54.4$ & $64.9$ & $74.1$ & $77.8$ & -
& $16.6$ & $39.1$ & $56.3$ & $77.5$ & $83.9$ & -
& $57.5$ & $61.3$ & $64.2$ & $\mathbf{79.4}$ & $69.1$ & -
& $71.3$ & $71.7$ & $71.9$ & $69.9$ & $69.1$ & - \\

\textbf{Forgetting}
& $22.4$ & $49.8$ & $63.3$ & $74.3$ & $78.1$ & -
& $11.8$ & $37.0$ & $56.7$ & $72.2$ & $81.1$ & -
& $31.6$ & $43.6$ & $60.1$ & $73.4$ & $77.0$ & -
& $61.1$ & $68.5$ & $67.9$ & $\underline{74.3}$ & $\underline{74.3}$ & - \\

\textbf{EL2N}
& $16.8$ & $44.4$ & $61.2$ & $74.4$ & $\mathbf{78.5}$ & -
& $8.9$ & $28.4$ & $49.2$ & $71.0$ & $79.3$ & -
& $31.0$ & $52.3$ & $64.2$ & $72.9$ & $76.5$ & -
& $\mathbf{74.6}$ & $\mathbf{74.3}$ & $\underline{75.6}$ & $61.0$ & $68.7$ & - \\

\textbf{Moderate}
& $34.6$ & $56.0$ & $64.9$ & $73.7$ & $77.5$ & -
& $19.7$ & $45.9$ & $59.8$ & $77.3$ & $83.8$ & -
& $60.5$ & $\underline{68.3}$ & $68.0$ & $71.1$ & $71.9$ & -
& $70.4$ & $72.0$ & $72.7$ & $68.4$ & $67.3$ & - \\

\textbf{CCS}
& $\underline{39.7}$ & $\underline{57.7}$ & $65.5$ & $73.6$ & $77.6$ & -
& $23.6$ & $49.3$ & $62.7$ & $77.8$ & $\mathbf{84.6}$ & -
& $52.8$ & $62.0$ & $67.1$ & $69.9$ & $70.7$ & -
& $72.9$ & $73.4$ & $\mathbf{76.7}$ & $\mathbf{75.3}$ & $\mathbf{76.7}$ & - \\

\textbf{$\mathbf{D^2}$-Pruning}
& $25.8$ & $54.8$ & $\underline{66.6}$ & $\underline{74.7}$ & $77.9$ & -
& $22.6$ & $\underline{50.9}$ & $\underline{63.9}$ & $72.4$ & $74.1$ & -
& $47.0$ & $49.2$ & $56.4$ & $60.8$ & $66.2$ & -
& $54.6$ & $61.0$ & $60.4$ & $64.3$ & $61.3$ & - \\

\textbf{InfoMax}
& $22.5$ & $52.0$ & $64.7$ & $72.7$ & $76.3$ & -
& $11.1$ & $37.3$ & $57.3$ & $75.5$ & $81.3$ & -
& $22.6$ & $54.0$ & $61.7$ & $70.9$ & $72.7$ & -
& $66.7$ & $62.1$ & $61.2$ & $62.3$ & $64.0$ & - \\

\textbf{RAM-APL}
& $39.3$ & $54.7$ & $61.7$ & $69.8$ & $74.8$ & -
& $\mathbf{26.0}$ & $\underline{50.9}$ & $61.5$ & $73.9$ & $81.0$ & -
& $30.2$ & $38.8$ & $51.2$ & $65.0$ & $68.2$ & -
& $45.5$ & $46.0$ & $47.2$ & $51.1$ & $53.3$ & - \\

\textbf{SDM}
& $37.5$ & $56.3$ & $64.5$ & $73.7$ & $78.2$ & -
& $20.6$ & $45.4$ & $62.5$ & $76.8$ & $82.4$ & -
& $\underline{61.0}$ & $64.3$ & $\underline{70.4}$ & $75.2$ & $74.4$ & -
& $71.4$ & $71.8$ & $70.9$ & $72.0$ & $74.0$ & - \\

\midrule[0.1pt]

\textbf{Ours}
& $\mathbf{42.9}$ & $\mathbf{59.4}$ & $\mathbf{67.4}$ & $\mathbf{75.2}$ & $\underline{78.3}$ & -
& $\underline{23.9}$ & $\mathbf{53.4}$ & $\mathbf{65.3}$ & $\mathbf{78.8}$ & $\underline{84.0}$ & -
& $\mathbf{68.6}$ & $\mathbf{72.7}$ & $\mathbf{77.9}$ & $\underline{79.3}$ & $\mathbf{79.8}$ & -
& $\underline{74.3}$ & $\underline{73.5}$ & $74.7$ & $72.3$ & $73.8$ & - \\

\bottomrule[1pt]

\end{tabular}
}

\end{table}
\begin{table}[tp]
\caption{
\textbf{Results on multimodal pretraining.}
Comparison of subset selection performance on CC12M using CLIP ViT-B/16.
We report zero-shot and linear-probe classification accuracy on ImageNet-1K, and Recall@1 for image-to-text (I2T) and text-to-image (T2I) retrieval on Flickr30K.
\textbf{Bold} and \underline{underlined} entries indicate the best and second-best results, respectively.
}
\label{tab:mm_results}
\vspace{-0.2cm}
\small
\setlength{\tabcolsep}{2.0pt}
\centering
\resizebox{\linewidth}{!}{
\begin{tabular}{l|cccccc|cccccc|cccccc|cccccc}
\toprule[1pt]
\textbf{Methods}
&\multicolumn{6}{c|}{\textbf{Zero-Shot ImageNet-1K}}
&\multicolumn{6}{c|}{\textbf{Linear Probe ImageNet-1K}}
&\multicolumn{6}{c|}{\textbf{I2T Flickr30K}}
&\multicolumn{6}{c}{\textbf{T2I Flickr30K}} \\

\cmidrule(lr){2-7}
\cmidrule(lr){8-13}
\cmidrule(lr){14-19}
\cmidrule(lr){20-25}

\textbf{Selection Rate}
&\textbf{10\%}&\textbf{20\%}&\textbf{30\%}&\textbf{50\%}&\textbf{70\%}&\textbf{100\%}
&\textbf{10\%}&\textbf{20\%}&\textbf{30\%}&\textbf{50\%}&\textbf{70\%}&\textbf{100\%}
&\textbf{10\%}&\textbf{20\%}&\textbf{30\%}&\textbf{50\%}&\textbf{70\%}&\textbf{100\%}
&\textbf{10\%}&\textbf{20\%}&\textbf{30\%}&\textbf{50\%}&\textbf{70\%}&\textbf{100\%} \\

\midrule[1pt]
\rowcolor{gray!15}
\textbf{Random}
&$3.6$&$7.7$&$12.9$&$20.9$&$26.4$&$32.4$
&$21.6$&$30.1$&$39.1$&$48.1$&$54.2$&$58.7$
&$\underline{3.8}$&$8.5$&$17.2$&$31.1$&$42.5$&$51.1$
&$\mathbf{3.1}$&$6.8$&$11.4$&$22.4$&$30.5$&$38.1$ \\

\midrule[0.1pt]
\textbf{CLIP Score}
&$3.2$&$9.9$&$15.2$&$23.9$&$28.6$&-
&$23.0$&$36.1$&$40.6$&$49.4$&$55.0$&-
&$\mathbf{3.9}$&$9.3$&$14.7$&$28.7$&$42.0$&-
&$2.7$&$7.0$&$10.6$&$20.1$&$31.0$&- \\

\textbf{K-center}
&$1.7$&$6.8$&$10.7$&$21.5$&$29.3$&-
&$18.5$&$31.8$&$38.5$&$49.6$&$55.8$&-
&$1.8$&$12.6$&$21.3$&$\mathbf{38.4}$&$\underline{46.4}$&-
&$1.6$&$8.3$&$13.9$&$\underline{26.7}$&$\underline{33.8}$&- \\

\textbf{Moderate}
&$3.8$&$10.1$&$14.6$&$22.5$&$29.0$&-
&$22.3$&$34.9$&$41.7$&$50.1$&$55.8$&-
&$3.1$&$11.0$&$19.1$&$32.8$&$43.3$&-
&$2.4$&$8.3$&$13.0$&$22.9$&$32.4$&- \\

\textbf{CCS}
&$\underline{4.0}$&$\underline{10.3}$&$\underline{15.4}$&$22.5$&$25.2$&-
&$\underline{23.6}$&$\underline{36.3}$&$42.5$&$50.0$&$53.5$&-
&$3.7$&$\underline{13.5}$&$\underline{22.3}$&$35.4$&$41.7$&-
&$\underline{2.8}$&$\underline{10.0}$&$\underline{15.4}$&$24.7$&$29.3$&- \\

\textbf{$\mathbf{D^2}$-Pruning}
&$2.5$&$8.8$&$14.4$&$\underline{24.3}$&$\underline{29.6}$&-
&$22.3$&$36.0$&$\underline{42.7}$&$\underline{51.7}$&$\underline{56.6}$&-
&$2.5$&$9.6$&$20.0$&$33.5$&$44.9$&-
&$2.0$&$7.6$&$13.1$&$23.2$&$31.4$&- \\

\textbf{Infomax}
&$2.3$&$8.9$&$14.2$&$24.1$&$29.2$&-
&$22.1$&$36.1$&$41.4$&$51.2$&$55.9$&-
&$2.3$&$9.4$&$19.3$&$34.1$&$43.8$&-
&$2.1$&$7.3$&$12.7$&$23.5$&$30.8$&- \\

\midrule[0.1pt]
\textbf{Ours}
&$\mathbf{6.4}$&$\mathbf{12.4}$&$\mathbf{16.6}$&$\mathbf{25.9}$&$\mathbf{30.9}$&-
&$\mathbf{24.7}$&$\mathbf{37.0}$&$\mathbf{43.9}$&$\mathbf{52.8}$&$\mathbf{57.3}$&-
&$3.7$&$\mathbf{15.4}$&$\mathbf{24.7}$&$\underline{37.2}$&$\mathbf{48.7}$&-
&$2.6$&$\mathbf{10.2}$&$\mathbf{16.5}$&$\mathbf{27.6}$&$\mathbf{35.8}$&- \\

\bottomrule[1pt]
\end{tabular}
}
\vspace{-0.45cm}
\end{table}

\subsection{Main Results}

\mypara{Image classification}
We evaluate subset selection across fine-grained recognition and spurious-correlation benchmarks in Table~\ref{tab:cls_results}.
Our method shows its largest advantages in the low-data regime.
On NABirds, it improves over the strongest baseline by $3.2\%$ at the $10$\% selection rate, with consistent gains as the budget increases.
The advantage is even more pronounced on Waterbirds, reaching $7.6\%$ and $7.5\%$ at the $10$\% and $30$\% rates, respectively, indicating substantially better robustness.
Our method also achieves the best performance at multiple budgets on Stanford Cars while remaining competitive on Spawrious. EL2N and CCS yield better performance on Spawrious, but their performance on other benchmarks is generally weak. 

\mypara{Multimodal pretraining}
We further evaluate whether the same selection principle transfers to large-scale multimodal pretraining by selecting CC12M subsets and training CLIP ViT-B/16 (Table~\ref{tab:mm_results}).
Our method selects data samples based on the image-level concept support. Even \textit{without information from the text distribution}, our method consistently achieves the best ImageNet-1K zero-shot and linear-probe accuracy across selection rates, with a $2.4\%$ improvement in zero-shot accuracy at the $10$\% budget.
The benefit also extends beyond classification to cross-modal retrieval: at $30$\%, I2T and T2I Recall@1 improve by $2.4\%$ and $1.1\%$ over the strongest baselines, and the corresponding margins are $2.3\%$ and $2.0\%$ at $70$\%.
These results indicate that the concept-support principle transfers beyond image classification and improves image--text alignment at multimodal pretraining scale.

\subsection{Ablation Study}\label{sec:ablation}

\begin{figure}[t]
\begin{minipage}[t]{0.65\textwidth}
    \centering
    \includegraphics[width=\linewidth]{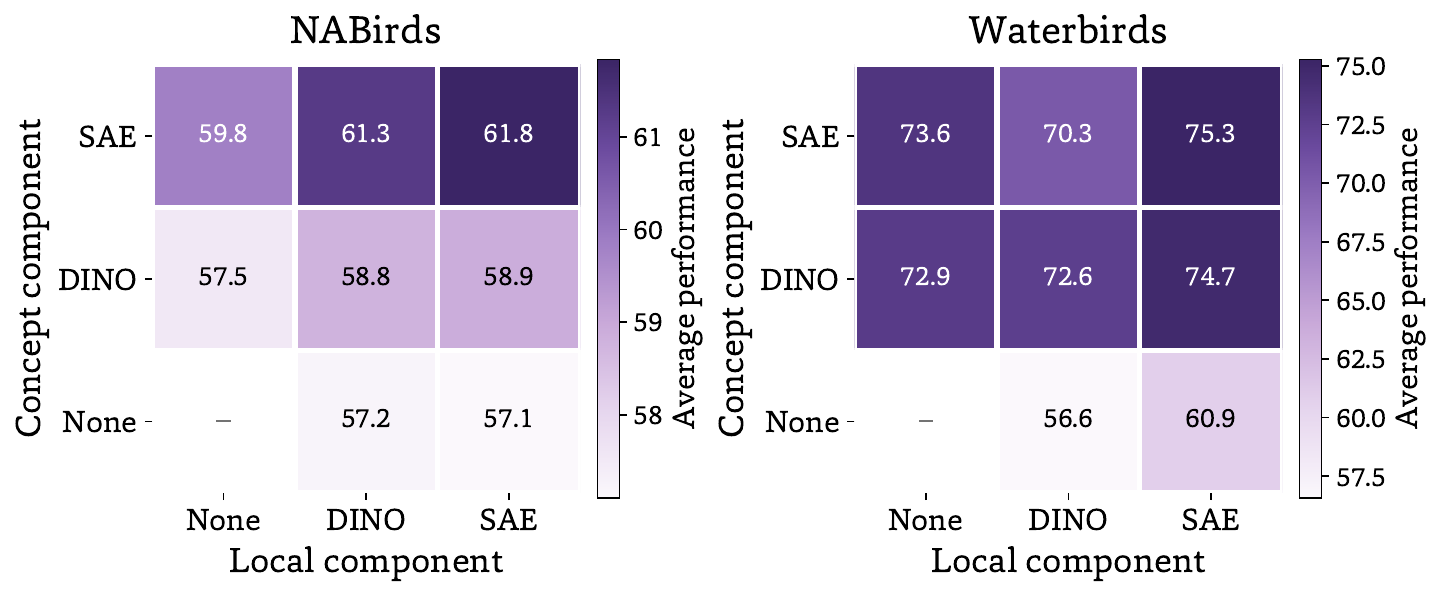}
    \vspace{-12pt}
    \caption{
    \textbf{Ablation of representations used for concept coverage and local support.}
    Rows indicate the representation used for the concept component, and columns indicate the representation used for the local component; \emph{None} removes the corresponding component.
    We report NABirds classification accuracy and Waterbirds worst-group accuracy (WGA), averaged across the 10\%, 30\%, and 50\% selection rates.
    }
    \label{fig:feature_space_ablation}
\end{minipage}
\hfill
\begin{minipage}[t]{0.32\textwidth}
    \centering
 \includegraphics[width=\linewidth]{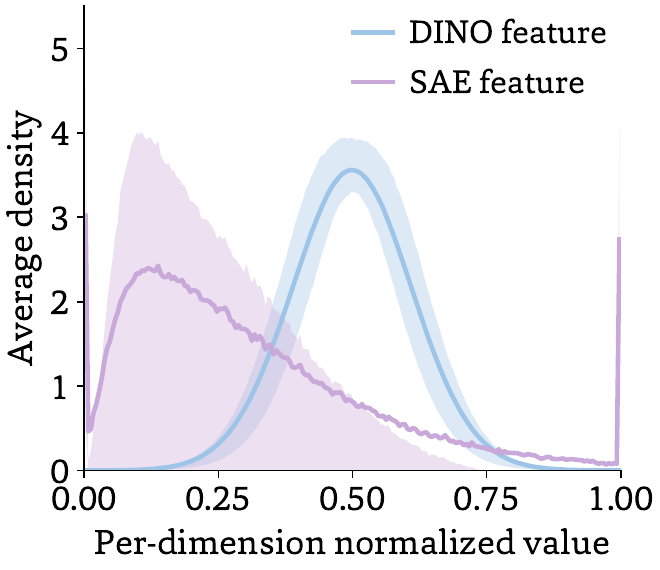}
    \vspace{-12pt}
    \caption{
    \textbf{Activation distributions of DINO and SAE features.}
    We normalize each DINO feature dimension and SAE feature dimension independently, and report the average activation density across dimensions.
    }
    \label{fig:dino_vs_sae_distribution}
\end{minipage}
\vspace{-4pt}
\end{figure}

\begin{table}[t]
\centering
\small
\setlength{\tabcolsep}{2.5pt}
\caption{\textbf{(a) Robustness to different pretrained features} used for SAE training and subset selection. \textbf{(b) Effect of normalization} over neighbor similarity.
We report accuracy for NABirds and worst-group accuracy (WGA) for Waterbirds averaged across selection ratios.
Best results are \textbf{bolded}.}
\vspace{-4pt}
\begin{subtable}{0.6\textwidth}
\caption{}
\label{tab:feature_robustness}
\resizebox{\textwidth}{!}{
\begin{tabular}{@{}l ccccc@{}}
\toprule
\multirow{2}{*}{\textbf{Dataset}} & \multicolumn{5}{c}{\textbf{Pretrained Architecture}} \\
& \textbf{DINOv2-B}
& \textbf{DINOv3-S}
& \textbf{DINOv3-B}
& \textbf{CLIP ViT-B/16}
& \textbf{SigLIP 2} \\
\midrule

NABirds Acc
& $\mathbf{62.0}$ & $\mathbf{62.0}$ & $61.8$ & $61.6$ & $60.8$ \\

Waterbirds WGA
& $75.1$ & $\mathbf{76.8}$ & $75.3$ & $76.0$ & $75.7$ \\

\bottomrule
\end{tabular}
}
\end{subtable}
\hfill
\begin{subtable}{0.38\textwidth}
\setlength{\tabcolsep}{6pt}
\caption{}\label{tab:row_normalization}
\resizebox{\textwidth}{!}{
\begin{tabular}{@{}l cc@{}}
\toprule
\multirow{2}{*}{\textbf{Dataset}} & \multicolumn{2}{c}{\textbf{Local Support}} \\
& \textbf{Unnormalized}
& \textbf{Normalized} \\
\midrule

NABirds Acc
& $61.5$ & $\mathbf{61.8}$ \\

Waterbirds WGA
& $74.6$ & $\mathbf{75.3}$ \\

\bottomrule
\end{tabular}
}
\end{subtable}






\vspace{-8pt}

\end{table}

\mypara{Concept and local representations}
Our proposed objective combines  concept coverage and local support. Both metrics can be calculated from dense embeddings or sparse concept activation vectors. We ablate this representation usage in  Figure~\ref{fig:feature_space_ablation}. 
On NABirds, using SAE features for concept coverage consistently provides the strongest results. On Waterbirds, we see a stronger performance improvement when moving the calculation of local support from dense DINO embeddings to the SAE space. 
We further investigate why the sparse concepts identified by SAEs facilitate better selection despite using the same objective by comparing their dimensional distribution across the dataset. As shown in Figure~\ref{fig:dino_vs_sae_distribution}, DINO features generally illustrate normal distributions across samples. In contrast, SAEs decompose dense features into monosemantic units and yield multiple peaks per dimension. Such distributions create clear separation between samples with different characteristics. Thus, the data selection objective receives more indicative signals for the accumulated and rare concepts.

\mypara{Robustness to pretrained visual features}
Our method is not tied to a particular pretrained representation. In
Table~\ref{tab:feature_robustness}, we construct the concept representation from different
pretrained visual encoders, including DINOv2-B~\citep{oquab2024dinov},
DINOv3-S and DINOv3-B, CLIP-B,
and SigLIP 2-B~\citep{tschannen2025siglip2}.
The method remains consistently effective across these substantially different feature spaces,
demonstrating that its performance does not depend on a specific backbone or pretraining
objective. This robustness aligns with our formulation: the selection objective examines the accumulated distribution to locate tail concepts. Therefore, any encoder that extracts diverse concepts can provide sufficient information for our approach. Thus, the concept-based subset
selection can serve as a general selection principle on top of different pretrained visual features.

\begin{figure}[t]

%
\begin{minipage}{0.59\textwidth}
\begin{subfigure}{0.48\textwidth}
    \centering
    \includegraphics[width=\textwidth]{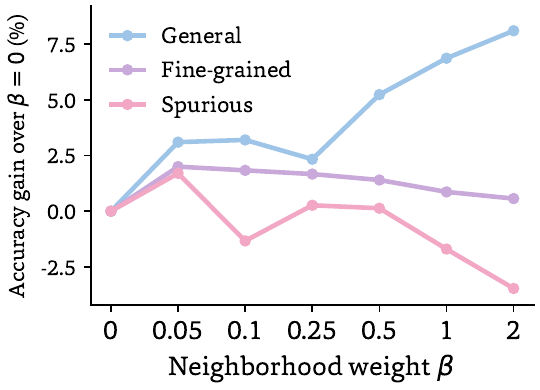}
    \caption{}\label{fig:beta_sweep}
\end{subfigure}
\hfill
\begin{subfigure}{0.5\textwidth}
    \centering
    \raisebox{2pt}{
    \includegraphics[width=\textwidth]{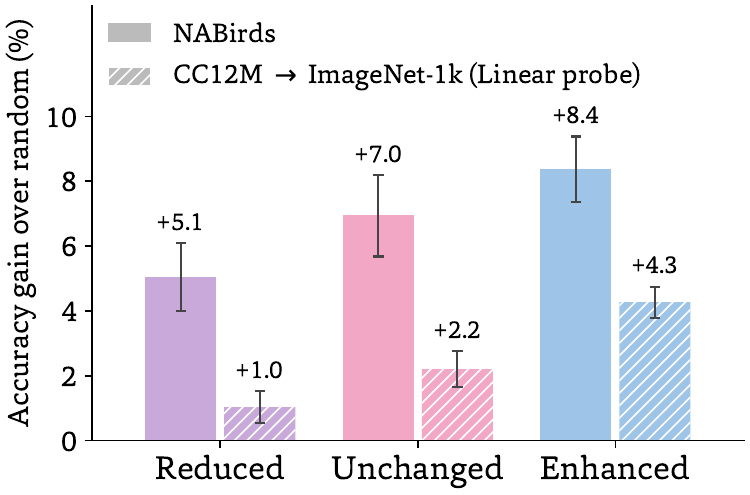}
    }
    \caption{}\label{fig:group-gain-slope}
\end{subfigure}
    \vspace{-2pt}
    \caption{
    \textbf{(a) Sensitivity to the neighborhood weight $\beta$}, averaged over the 10\%, 30\%, and 50\% selection rates, for general (Tiny-ImageNet), fine-grained (NABirds), and spurious-correlation (Waterbirds, WGA) settings.
    \textbf{(b) Concept support and downstream gain.}
    Accuracy gain over random selection on images containing concepts with reduced, unchanged, or enhanced training support.}

\end{minipage}
\hfill
\begin{minipage}{0.39\textwidth}

    \captionsetup{type=table}
    \caption{
    \textbf{Concept-support profiles of NABirds subsets.}
    Unique Coverage: the number of concepts receiving nonzero support.
    Effective Number: exponential of the entropy of the normalized concept-mass distribution.
    Prevalence Ratio: mass-weighted mean concept prevalence normalized by the full set; values below one indicate greater relative support for tail concepts.
    }
    \label{tab:concept-profile}

    \centering
    \small
    \setlength{\tabcolsep}{5pt}
    \vspace{-4pt}

    \resizebox{\textwidth}{!}{
    \begin{tabular}{@{}l|ccc@{}}
        \toprule

        \textbf{Methods}
        & \shortstack{\textbf{Unique}\\\textbf{Coverage}}
        & \shortstack{\textbf{Effective}\\\textbf{Number}}
        & \shortstack{\textbf{Prevalence}\\\textbf{Ratio}} \\

        \midrule

        \textbf{Random}
        & $7{,}973$
        & $865$
        & $1.00$ \\

        \midrule

        \textbf{K-Center}
        & $8{,}215$
        & $893$
        & $0.97$ \\

        \textbf{Forgetting}
        & $8{,}120$
        & $888$
        & $0.97$ \\

        \textbf{D$^2$-Pruning}
        & $8{,}099$
        & $887$
        & $0.97$ \\

        \midrule

        \textbf{Ours}
        & $\mathbf{8{,}951}$
        & $\mathbf{1{,}124}$
        & $\mathbf{0.74}$ \\

        \bottomrule
    \end{tabular}
    }

\end{minipage}
\vspace{-8pt}
\end{figure}

\mypara{Effect of similarity normalization}
We normalize the similarity between each sample and its neighbors to produce the local support component in Eq.~\ref{eq:local}. We examine its necessity in Table~\ref{tab:row_normalization}. Specifically, we compare directly using raw cosine similarities with
normalized ones. Normalization yields slightly better performance across selection
rates. It gives each target sample a bounded contribution budget that is
distributed among its neighbors according to their relative similarities, preventing samples in
dense neighborhoods from contributing disproportionately large local support. 

\mypara{Sensitivity to neighborhood weight}
Finally, we study the neighborhood weight $\beta$, which controls the relative contribution of
local support to the selection objective, in Figure~\ref{fig:beta_sweep}.
Setting $\beta=0$ removes the local contribution and simplifies the objective to concept coverage alone.
Particularly, we test its influence on different types of original data.

Across fine-grained, spurious, and general cases, having $\beta=0.05$ consistently improves the performance over concept coverage alone. However, their preferred weights differ. Spurious correlation and fine-grained datasets yield the best training performance at $\beta=0.05$, whereas
general-purpose datasets favor a substantially stronger local contribution with $\beta=2$.
This suggests that when the task scale is relatively small and concentrated, broader concept coverage reduces the risk of test-time distribution shift. In contrast, general classification tasks, which usually have larger concept breadth, benefit from emphasizing local support. We therefore use the corresponding optimal value for each dataset group throughout our experiments. See \S\ref{sec:obj_design} for additional ablations and per-rate results.

\section{Analysis and Discussion}\label{sec:analysis}
The concept allocation produced by data selection creates a direct connection to downstream model behavior. We examine the concept distribution under different selection strategies and how model behavior changes accordingly, asking whether the evidence made more available during training is also where downstream recognition improves. 

\mypara{Different samples can produce similar concept support}
\label{sec:concept-shift}
Selecting images with different criteria need not substantially change the concept distribution in the resulting subset. We summarize the concept coverage produced by K-Center, Forgetting, D$^2$-Pruning, and our approach in Table~\ref{tab:concept-profile}. 
\textbf{Unique Coverage} counts concepts receiving nonzero support; \textbf{Effective Number} measures how broadly normalized activation mass is distributed; and \textbf{Prevalence Ratio} measures its mass-weighted mean concept prevalence relative to the full dataset.
A prevalence ratio below one indicates greater relative support for tail concepts. Despite distinct selection criteria, their allocations of activation mass across concepts are similar, as confirmed in Figure~\ref{fig:teaser}.
Our subset supports $8{,}951$ concepts, compared with $7{,}973$ under random selection, and increases the effective number from $865$ to $1{,}124$.
Its prevalence ratio falls from $1.00$ to $0.74$, while the three compared selectors remain at $0.97$. The enrichment profile in Figure~\ref{fig:teaser} shows our subset allocates a greater share of its concept support to tail concepts.
The change proves that our approach effectively reallocates the concept distribution compared with the original dataset and other baselines.

\mypara{From concept support to better recognition}
\label{sec:concept-gain}
The change in concept support motivates us to investigate their downstream correspondence. We start with test accuracy and use NABirds and ImageNet-1K linear probing from multimodal pretraining as test tasks. 
Based on the change in its training support compared with random selection, we label each concept as \textbf{enhanced}, \textbf{unchanged}, or \textbf{reduced}. For each concept, we select the test samples with the highest activation and calculate their average accuracy. Then, we measure how much our model improves over the random-selection model on each concept category. As shown in Figure~\ref{fig:group-gain-slope}, while all three categories improve, samples carrying enhanced concepts show substantially larger improvements than the other two categories.
The observation proves that concepts enhanced by our approach, while remaining in the distributional tail, actively shape model behavior.
It also raises another question: \textit{do these enhanced concepts always relate to specific categories}? We answer this question using two case studies.

\begin{figure}[t]
    \centering
    \includegraphics[width=\linewidth]{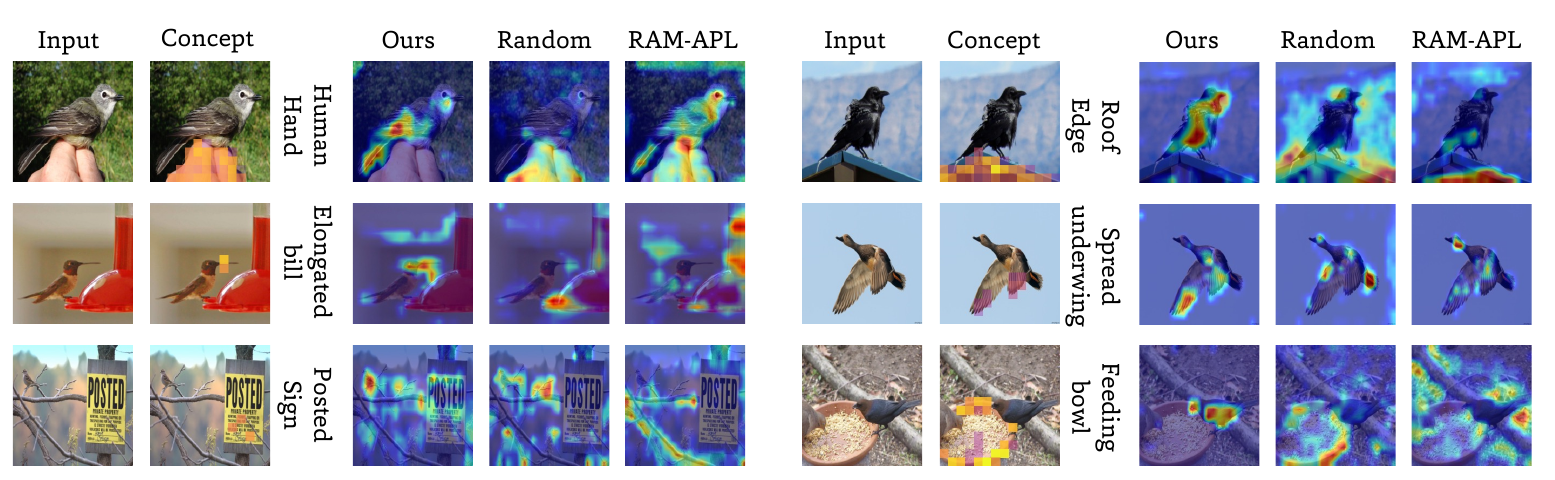}
    \vspace{-16pt}
    \caption{
    \textbf{Qualitative examples of enhanced visual concepts.}
    Our selection enhances concepts of both fine-grained attributes and contextual evidence.
    Finer-CAM visualizations compare the regions that models trained on our, random, and RAM-APL subsets rely on for prediction, showing that broader concept exposure during training improves test-time robustness.
    }
    \label{fig:concept_finer_cam}
    \vspace{-12pt}
\end{figure}

\subsection{Case Study: NABirds}
We first look into fine-grained classification and show example concepts in NABirds with enhanced support produced by our selection approach in Figure~\ref{fig:concept_finer_cam}.
Some concepts are activated on discriminative features, such as beaks and wings. At the same time, other concepts that do not directly relate to categorical appearance, such as hands, signs, and feeders, are also enhanced. Test accuracy on samples carrying these concepts also improves.

\mypara{Beyond category-related concepts}
Using the concept \textit{hand} as an example, we examine its support over the dataset. Among $23{,}929$ training images, only $160$ activate the dimension, spread across $108$ categories. Random selection retains $20$ such images across $20$ categories, whereas our subset retains $60$ across $50$ categories, substantially broadening the support to this uncommon context. On strongly hand-activated test images, accuracy rises from $28.6\%$ to $51.4\%$. We visualize the explanation heatmap using Finer-CAM~\citep{zhang2025finer} to understand this behavior, which suggests that the model trained on our selected data remains concentrated on the bird, while data selected by random or RAM-APL biases the model to these contextual concepts.
These analyses show that while tail contextual concepts are not category-related discriminative features, they increase the robustness of model training, so these concepts remain in distribution at test time.
Broader category coverage also reduces the risk of \textit{hand} spuriously correlating with a small number of categories.

\mypara{Necessity of local support}
\label{sec:tail-limits}
While some tail concepts improve training robustness, not all do. In Figure~\ref{fig:tail-limits}, we show that removing the local support term only slightly increases the amplification on tail concepts, but reduces accuracy from $42.9\%$ to $37.9\%$. If we rank samples only by their concept rarity, which is the inverse prevalence of their active concepts, the tail concepts are substantially enriched by $8.2\times$, but the resulting model only yields $33.1\%$ accuracy. Aggressively prioritizing tail concepts can discard useful representative structure. 
Figure~\ref{fig:rarity-neighbors} partially explains that pure-rarity selection often favors atypical examples with little local support, whereas our objective favors candidates that also represent under-supported evidence within their local
neighborhoods.

\subsection{Case Study: Waterbirds}
After locating category-irrelevant concepts that contribute to model robustness, we run another test to expose how our approach alters support for spurious correlations. Waterbirds provides a controlled setting in which bird identity is spuriously correlated with background. 
 
\begin{figure}[t]
%
\begin{minipage}[t]{0.58\textwidth}
\begin{subfigure}[t]{0.58\linewidth}
    \vspace{0pt}
    \centering
    \includegraphics[width=\textwidth]{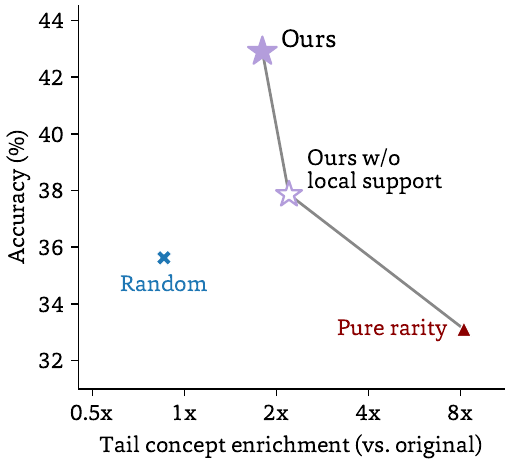}
    \vspace{-5pt}
    \caption{}
    \label{fig:tail-limits}
\end{subfigure}
\hfill
\begin{subfigure}[t]{0.40\linewidth}
    \vspace{0pt}
    \centering
    \includegraphics[width=\textwidth]{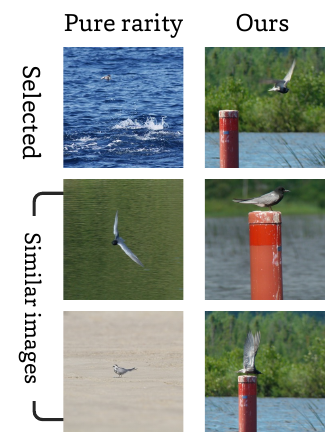}
    \vspace{-5pt}
    \caption{}
    \label{fig:rarity-neighbors}
\end{subfigure}

\vspace{-7pt}

\caption{
\textbf{Local support identifies useful tail examples without overemphasizing rarity.}
\textbf{(a)} Further increasing tail enrichment beyond the full objective does
not improve NABirds accuracy, indicating that rarity alone is insufficient.
\textbf{(b)} Pure-rarity selection can retain isolated examples, whereas our
local-support objective favors underrepresented examples that remain supported
by nearby samples.
}
\label{fig:rarity-tail}

\end{minipage}
\hfill
\begin{minipage}[t]{0.40\textwidth}
\vspace{0pt}

\captionsetup{type=table}
\caption{
\textbf{Subset composition and robustness to background changes on Waterbirds.}
Counter-corr.\ is the fraction of bias-conflicting examples.
All-bg correct requires correct prediction under all four background replacements.
Group-matched controls use the same label--background counts as Ours.
All values are percentages.
}
\label{tab:waterbirds-shortcut}

\centering
\small
\setlength{\tabcolsep}{3.0pt}
\vspace{-4pt}

\resizebox{\textwidth}{!}{
\begin{tabular}{@{}lccc@{}}
\toprule[1pt]

\textbf{Method}
& \textbf{Counter-corr.}
& \textbf{WGA} $\uparrow$
& \textbf{All-bg} $\uparrow$ \\

\midrule

\textbf{Random}
& $4.0$
& $52.3$
& $57.6$ \\

\midrule

\textbf{Forgetting}
& $5.0$
& $31.6$
& $32.4$ \\

\textbf{K-Center}
& $5.4$
& $57.5$
& $72.6$ \\

\textbf{Moderate}
& $9.3$
& $\underline{60.5}$
& $78.3$ \\

\midrule

\textbf{Ours}
& $18.8$
& $\mathbf{68.6}$
& $\mathbf{86.5}$ \\

\midrule

\multicolumn{4}{l}{\textit{Group-matched controls}} \\

\midrule

\textbf{Random}
& $18.8$
& $63.7$
& $84.2$ \\

\textbf{Difficult}
& $18.8$
& $52.2$
& $73.7$ \\

\textbf{Easy}
& $18.8$
& $64.0$
& $\underline{85.8}$ \\

\bottomrule

\end{tabular}
}

\end{minipage}

\vspace{-8pt}
\end{figure}
\mypara{Selection and background reliance}
\label{sec:background-reliance}
As shown in Table~\ref{tab:waterbirds-shortcut}, our approach yields $18.8\%$ bias-conflicting examples (\textit{e.g.}, waterbirds with land background) in the selected set, while the other compared approaches select less than $10\%$. 
Although this ratio does not have a strict positive relationship with downstream performance, broader concept coverage weakens the background shortcut by diversifying the object--background combinations.

\mypara{Group composition examination}
Part of this robustness improvement could arise from the larger fraction of bias-conflicting examples alone. We therefore construct random subsets with the same number of samples from four label--background groups. Matching the group composition improves WGA from $52.3\%$ for ordinary random selection to $63.7\%$, still below our method. 

Moreover, with the same group quotas, selecting difficult examples yields only $52.2\%$ WGA, while selecting easy examples reaches $64.0\%$. These results are consistent with prior evidence that group proportions alone do not determine coreset robustness~\citep{dharmasiri2025impact}. Together, the results suggest that concept-based selection improves robustness through both a more favorable group composition and richer visual evidence within those groups.

\section{Conclusion}
This paper views data selection as reallocating training support for visual concepts under a limited data budget and studies the relationship between allocation strategy and downstream model behavior. Specifically, we extract visual concepts from each image using sparse autoencoders. Each selected sample contributes its concepts to an accumulated distribution throughout the selection process. We design an objective that balances concept coverage and local support to mitigate the gap in tail concepts while avoiding isolated samples. This flexibility also allows us to vary allocation strategies across fine-grained, spurious-correlation, and general selection scenarios. Experiments on classification and multimodal pretraining tasks suggest the effectiveness of the proposed concept-based selection. Importantly, we use the concepts enhanced by our method to probe model properties trained on the selected dataset. Analyses show that context concepts, while not directly contributing category-related information, improve model robustness when encountering out-of-distribution and spurious connections. More broadly, this paper advocates for concepts, which offer interpretable, fine-grained quantitative measurement for visual semantics, as an important principle in data selection. We need selection approaches that can proactively shape downstream model behavior.

\section*{Ethics statement}

Our study does not involve human subjects, and no personal or identifiable
information was collected. All experiments are conducted on publicly
available datasets in accordance with their respective licenses. We
emphasize that the released coresets and models are intended for scientific
research purposes. Since our method selects subsets from existing datasets,
the selected data naturally reflect the characteristics of their source
datasets. We therefore recommend validating the selected subsets for the
intended application before use beyond research settings.

\section*{Reproducibility statement}
We release our source code, the selected coresets, and the trained
models on our \href{https://imageomics.github.io/mind-the-tail/}{project website}.
The experimental setup and training hyperparameters are presented
in \S\ref{sec:exp_details}.
The details of SAE training and evaluation are included in
\S\ref{sec:sae_appendix}.
Efficient implementation at scale is described in
\S\ref{sec:efficient}.

\section*{Acknowledgments}

Our research is supported by NSF OAC~2118240 and resources from the Ohio Supercomputer Center~\citep{OhioSupercomputerCenter1987}. The authors are grateful for the generous support of the computational resources from the Ohio Supercomputer Center.


\bibliography{references}
\bibliographystyle{preprint}


\newpage
\appendix

\DoToC

\newpage

\section{Implementation Details}\label{sec:exp_details}

\subsection{Default Configuration}
We train a TopK SAE~\citep{makhzani2014ksparse} with Matryoshka objective  \S\ref{sec:sae_obj} on layer-11 patch-token
activations extracted from a frozen DINOv3-B encoder~\citep{simeoni2025dinov3}. We use $100$ million patch tokens sampled from
ImageNet-1K~\citep{russakovsky2015imagenet} for training and $10$ million patch tokens for validation. The SAE has $16{,}384$ latent dimensions, with $k=32$ active latents per token. After training, we run SAE inference on each candidate dataset and max-pool the patch-level activations to obtain one concept vector per image, on which data selection is performed.
We use the logarithmic utility with $\beta=0.05$ for fine-grained and spurious-correlation settings, $\beta=2$ for general setting and $k=3$ nearest neighbors by default.
For classification, ResNet-50~\citep{he2016deep} is initialized from ImageNet-pretrained weights, except on Tiny-ImageNet and ImageNet-$1$K, where it is trained from scratch.
For multimodal pretraining, a ViT-B/16 is trained from scratch on the selected CC12M subsets.
All selection methods share the same training configuration for each task.

\subsection{Training Hyperparameters}

We summarize the hyper-parameter configurations in \autoref{tab:hyper-imagenet} and \autoref{tab:hyper-classification} for classification, \autoref{tab:hyper-cc12m} for multimodal pretraining, and \autoref{tab:hyper-sae} for SAE training. ImageNet-1K uses a larger learning rate, a cosine schedule, and a longer training budget due to its scale. The number of training steps is the same for every selection rate.

\subsection{Scalable Selection: Neighbor Graph and Lazy Greedy}\label{sec:efficient}
\mypara{Neighbor graph}
On CC12M, we build the $k$-nearest-neighbor graph in SAE space
with approximate search. Sparse concept vectors are randomly projected to
$512$ dimensions~\citep{achlioptas2003database}, candidate neighbors are retrieved
with a FAISS inverted-file index with product
quantization~\citep{jegou2011product,johnson2019billion}, and the candidates are
re-ranked by exact cosine similarity on the original sparse vectors before
keeping the top $k$. The resulting neighborhood weights are
materialized once as a sparse matrix, so that $\tilde{x}_i$ is computed a
single time for every image. On a single CPU node with $56$ threads, candidate
retrieval takes $26$ minutes and the exact re-ranking $8$ minutes, so the graph
for the 12M-image pool is ready in about $34$ minutes.

\mypara{Global lazy greedy}
Selection runs as one lazy greedy pass~\citep{minoux1978accelerated} over the
whole pool. Each marginal gain touches only the nonzero
entries of $\tilde{x}_i$, and the lazy evaluation refreshes about $50$ candidates
per selected image on average. The greedy pass takes
$47$ minutes on the same node, so the full ranking of CC12M costs about
$1.3$ hours in total and is reused for all ratios.

\begin{table*}[t]
\centering


\begin{minipage}[t]{0.48\textwidth}
\centering
\caption{The adopted hyper-parameter setting for image classification on ImageNet-1K.}
\label{tab:hyper-imagenet}

\small
\setlength{\tabcolsep}{4pt}
\renewcommand{\arraystretch}{1.05}

\begin{tabular}{@{}p{0.38\linewidth}p{0.56\linewidth}@{}}
\toprule
\textbf{Setting} & \textbf{Value} \\
\midrule
Model            & ResNet-50 \\
Initialization   & None \\
GPUs             & $1\times$ H100 \\
Batch size       & $256$ \\
Optimizer        & SGD \\
Learning rate    & $0.1$ \\
Momentum         & $0.9$ \\
Weight decay     & $1\times10^{-4}$ \\
Scheduler        & Cosine \\
Training steps   & $400{,}000$ \\
\bottomrule
\end{tabular}
\end{minipage}
\hfill
\begin{minipage}[t]{0.48\textwidth}
\centering
\caption{The adopted hyper-parameter setting for image classification on other datasets.}
\label{tab:hyper-classification}

\small
\setlength{\tabcolsep}{4pt}
\renewcommand{\arraystretch}{1.05}

\begin{tabular}{@{}p{0.38\linewidth}p{0.56\linewidth}@{}}
\toprule
\textbf{Setting} & \textbf{Value} \\
\midrule
Model            & ResNet-50 \\
Initialization   & ImageNet / None:Tiny-In1K\\
GPUs             & $1\times$ H100 \\
Batch size       & $256$ \\
Optimizer        & SGD \\
Learning rate    & $0.001$ \\
Momentum         & $0.9$ \\
Weight decay     & $1\times10^{-4}$ \\
Scheduler        & Constant \\
Training steps   & $18{,}700$ \\
\bottomrule
\end{tabular}
\end{minipage}

\vspace{0.6cm}

\begin{minipage}[t]{0.48\textwidth}
\centering
\caption{The adopted hyper-parameter setting for multimodal pretraining on CC12M.}
\label{tab:hyper-cc12m}

\small
\setlength{\tabcolsep}{4pt}
\renewcommand{\arraystretch}{1.05}

\begin{tabular}{@{}p{0.38\linewidth}p{0.56\linewidth}@{}}
\toprule
\textbf{Setting} & \textbf{Value} \\
\midrule
Model            & ViT-B/16 \\
Initialization   & None \\
GPUs             & $4\times$ H100 \\
Batch size       & $16{,}384$ \\
Optimizer        & AdamW \\
Learning rate    & $1\times10^{-3}$ \\
Weight decay     & $0.5$ \\
Scheduler        & Cosine decay \\
Warmup           & $670$ steps \\
Training steps   & $10{,}050$ \\
\bottomrule
\end{tabular}
\end{minipage}
\hfill
\begin{minipage}[t]{0.48\textwidth}
\centering
\caption{The adopted hyper-parameter setting for SAE training.}
\label{tab:hyper-sae}

\small
\setlength{\tabcolsep}{4pt}
\renewcommand{\arraystretch}{1.05}

\begin{tabular}{@{}p{0.43\linewidth}p{0.51\linewidth}@{}}
\toprule
\textbf{Setting} & \textbf{Value} \\
\midrule
Encoder layer        & Layer 11 patch tokens \\
SAE dictionary size  & $16{,}384$ \\
GPUs  & $1\times$ H100 \\
SAE activation       & Top-$K$ \\
Active latents/token & $k=32$ \\
Token batch size     & $16{,}384$ \\
Optimizer            & Adam \\
Learning rate        & $1\times10^{-3}$\\
Warmup               & $500$ steps \\
Training steps       & $6{,}104$ \\
\bottomrule
\end{tabular}
\end{minipage}

\end{table*}

\newpage

\section{Sparse Autoencoder Training and Evaluation}\label{sec:sae_appendix}

\subsection{Training Setup}\label{sec:sae_obj}

\mypara{SAE variants}
We train three sparse autoencoder (SAE) variants: ReLU, TopK, and BatchTopK SAEs
\citep{bricken2023monosemanticity,makhzani2014ksparse,bussmann2024batchtopk}.
Given an activation $\mathbf{x}\in\mathbb{R}^{d}$ extracted from a DINOv3-B encoder, each SAE computes
\[
\mathbf{z}
=
\phi\!\left(
\mathbf{W}_{\mathrm{enc}}\mathbf{x}
+
\mathbf{b}_{\mathrm{enc}}
\right),
\qquad
\hat{\mathbf{x}}
=
\mathbf{W}_{\mathrm{dec}}\mathbf{z}
+
\mathbf{b}_{\mathrm{dec}}.
\]
where $\phi$ determines the sparsification mechanism.

\mypara{Matryoshka training}
All variants are trained with the Matryoshka reconstruction objective
\citep{bussmann2025matryoshka}.
Given nested latent prefixes
$p_1 < \cdots < p_m = d_{\mathrm{SAE}}$, we reconstruct the input from each
prefix,
\[
\hat{\mathbf{x}}^{(p_i)}
=
\mathbf{W}_{\mathrm{dec}}[:,1:p_i]\,
\mathbf{z}_{1:p_i}
+
\mathbf{b}_{\mathrm{dec}}.
\]
and optimize
\[
\mathcal{L}_{\mathrm{SAE}}
=
\frac{1}{m}
\sum_{i=1}^{m}
\left\|
\mathbf{x}
-
\hat{\mathbf{x}}^{(p_i)}
\right\|_2^2
+
\lambda_{\mathrm{sp}}\mathcal{L}_{\mathrm{sp}}
+
\lambda_{\mathrm{aux}}\mathcal{L}_{\mathrm{aux}}.
\]
The nested objective encourages earlier latent dimensions to capture broadly
useful structure while later dimensions refine the reconstruction.
We use an $\ell_1$ sparsity penalty for the ReLU SAE.
For TopK and BatchTopK SAEs, sparsity is enforced directly by the TopK
activation, and we additionally use the AuxK loss
\citep{gao2025scaling} to mitigate dead latents.

\mypara{Training details}
We train SAEs on $100$ million patch-token activations from a DINOv3-B encoder.
All variants use ten Matryoshka prefixes; TopK and BatchTopK use $k=32$, and decoder columns are constrained to unit norm.
Full training hyperparameters are summarized in Table~\ref{tab:hyper-sae}.

\subsection{Evaluation Metrics and Results}
\mypara{Evaluation protocol}
We evaluate the learned ReLU, TopK, and BatchTopK SAEs on held-out activation
shards using complementary metrics that capture sparsity, reconstruction quality,
representation preservation, and downstream fidelity.

\mypara{$\boldsymbol{L}_0$}
$L_0$ denotes the mean number of active latent features per token.
It characterizes the effective sparsity of the SAE representation and is
interpreted jointly with reconstruction fidelity.

\mypara{Number of dead latents}
The number of dead latents, denoted $n_{\mathrm{dead}}$, counts latent features
that remain inactive or activate less frequently than a small threshold over
the evaluation set~\citep{bricken2023monosemanticity}.
Fewer dead latents indicate better utilization of the SAE dictionary.

\mypara{Normalized MSE}
Normalized mean squared error (MSE), also known as the fraction of variance
unexplained~\citep{gao2025scaling}, measures reconstruction fidelity by
normalizing the squared reconstruction error by the variance of the
mean-centered input.
A value of zero indicates perfect reconstruction, while a value of one
corresponds to predicting the mean activation.

\mypara{Cosine similarity (CS)}
Cosine similarity measures the directional agreement between each original
activation and its SAE reconstruction.
It provides a
complementary reconstruction metric to normalized MSE.
Higher values indicate better reconstruction fidelity.

\mypara{Centered kernel nearest-neighbor alignment (CKNNA)}
CKNNA~\citep{huh2024platonic} measures local representational alignment between
the original backbone activations and the corresponding SAE latent
representations.
Higher CKNNA indicates better preservation of the local geometry of the
original feature space.

\mypara{Downstream fidelity}
We replace the original patch-token activations at the trained layer with
their SAE reconstructions, propagate both representations through the remaining
frozen backbone layers, and apply the same fixed ImageNet-1K linear probe
~\citep{zaigrajew2025interpreting}.
We report the mean KL divergence between the original and reconstructed
prediction distributions together with the decrease in Top-1 accuracy.
Lower values indicate better preservation of the original model behavior.

\mypara{Results}
Table~\ref{tab:sae-evaluation} summarizes the quantitative comparison across
SAE variants.
TopK-32 provides a favorable balance across sparsity, reconstruction quality,
representation preservation, and downstream fidelity, and is therefore used in
our main experiments.

\begin{table}[!t]
\centering
\caption{\textbf{Quantitative comparison of SAE variants on ImageNet-1K.}
We compare ReLU SAEs with $\ell_1$ sparsity coefficients
$\lambda\in\{0.1,0.01\}$, TopK SAEs with $k\in\{16,32\}$ active latents per
token, and BatchTopK SAEs with an average batch-wide activation budget of
$k\in\{16,32\}$ latents per token. 
Arrows indicate the preferred direction.}
\label{tab:sae-evaluation}

\vspace{2pt}
\footnotesize
\setlength{\tabcolsep}{3.5pt}
\renewcommand{\arraystretch}{1.03}

\begin{tabular}{@{}lrrrrrrr@{}}
\toprule
\textbf{Run}
& $\mathbf{L}_0$
& \textbf{Norm. MSE} $\downarrow$
& $\mathbf{n}_{\mathrm{dead}}$ $\downarrow$
& \textbf{Mean KL} $\downarrow$
& \textbf{Top-1 Drop} $\downarrow$
& \textbf{CKNNA} $\uparrow$
& \textbf{CS} $\uparrow$ \\
\midrule

ReLU $\lambda=0.1$
& $6.75$ & $0.60$ & $6{,}331$ & $2.87$ & $0.51$ & $0.28$ & $0.91$ \\

ReLU $\lambda=0.01$
& $150.69$ & $0.22$ & $2{,}773$ & $0.24$ & $0.06$ & $0.46$ & $0.97$ \\

TopK-16
& $16.00$ & $0.32$ & $529$ & $0.34$ & $0.09$ & $0.59$ & $0.96$ \\

TopK-32
& $32.00$ & $0.26$ & $472$ & $0.21$ & $0.06$ & $0.69$ & $0.96$ \\

BatchTopK-16
& $16.04$ & $0.32$ & $230$ & $0.35$ & $0.09$ & $0.64$ & $0.96$ \\

BatchTopK-32
& $32.12$ & $0.26$ & $191$ & $0.22$ & $0.06$ & $0.52$ & $0.96$ \\

\bottomrule
\end{tabular}

\vspace{-0.5pt}
\end{table}

\newpage

\section{Analysis Protocols}\label{sec:analysis_protocols}

\subsection{Concept Support and Concept Groups}\label{sec:concept_group}
For each concept $c$, we measure its support in subset $S$ as
\[
m_c(S)=\sum_{I_i\in S} x_i[c].
\]
We draw $1{,}000$ random subsets using the same budgets
and use the resulting values of $m_c$ as a reference distribution.
A concept is \emph{enhanced} if its support under our selection exceeds the
$97.5$th percentile of this distribution, \emph{reduced} if it falls below the
$2.5$th percentile, and \emph{unchanged} if it lies within the $40$th--$60$th
percentiles.

\subsection{Pure-Rarity Selector}\label{sec:pure_rarity}

The pure-rarity selector isolates the effect of favoring rare concepts without
any coverage or local-support term. For each concept $c$, let $\rho_c$ be the
fraction of pool images on which it is active. Each image is scored by the
summed inverse prevalence of its active concepts,
\[
s_i
=
\sum_{c:\,x_i[c] > 0}
\frac{1}{\rho_c}.
\]
so that images carrying many rare concepts score highest.

\subsection{Visualization Methods}\label{sec:sae_vis}\label{sec:finercam_vis}

\mypara{SAE concept overlays}
Given an image $I$, we encode each of its DINOv3 patch tokens with the SAE,
giving patch-level concept activations $\mathbf{A}_{I}\in\mathbb{R}_{\geq 0}^{P\times C}$,
where $A_{I,p,c}$ denotes the activation of concept $c$ at patch $p$.
To visualize a concept $c$, we retrieve the images on which it is most strongly
active and show where in each image it fires.

Following \citet{stevens2025interpretable}, we arrange the activations of concept $c$
on the patch grid and normalize them by the maximum activation of $c$ over the
displayed images:
\begin{equation*}
\widehat{A}_{I,p,c}=\frac{A_{I,p,c}}{\max_{I',p'} A_{I',p',c}}.
\end{equation*}
Patches below a relative threshold are left untouched. The remaining patches are
colored by $\widehat{A}_{I,p,c}$ with a continuous colormap, with opacity
increasing with activation, and overlaid on the image. The overlay uses no gradients or labels.

\mypara{Finer-CAM attribution}
We adopt Finer-CAM~\citep{zhang2025finer} to visualize the regions that drive predictions of downstream ResNet-50 models trained on different selected subsets.
For each image, we backpropagate the logit of
its ground-truth category against the $3$ categories with the highest logits and apply Finer-CAM to the output of
the final convolutional stage (\texttt{layer4}).
The resulting heatmap is bilinearly upsampled to the input
resolution, normalized to $[0,1]$, and overlaid on the image.

\newpage

\section{Additional Ablations}\label{sec:obj_design}

\begin{table}[h]
\centering
\begin{minipage}[t]{0.49\textwidth}
\centering
\captionof{table}{\textbf{Ablation of representations used in concept coverage and local support.}
SAE and DINO denote representations derived from SAE
features and dense DINO features, respectively. We report accuracy for NABirds and worst-group accuracy (WGA) for Waterbirds. \textbf{Bold} and \underline{underlined} entries indicate the best and second-best results, respectively.}
\label{tab:feature_space_ablation}

\vspace{-2pt}

\small
\setlength{\tabcolsep}{4pt}
\resizebox{\textwidth}{!}{
\begin{tabular}{@{}ll ccc ccc@{}}
\toprule
\multicolumn{2}{c}{\textbf{Component}}
& \multicolumn{3}{c}{\textbf{NABirds}}
& \multicolumn{3}{c}{\textbf{Waterbirds}} \\
\cmidrule(lr){1-2}
\cmidrule(lr){3-5}
\cmidrule(lr){6-8}
Concept & Local
& 10\% & 30\% & 50\%
& 10\% & 30\% & 50\% \\
\midrule

SAE & SAE
& $\mathbf{42.9}$ & $\mathbf{67.4}$ & $\mathbf{75.2}$
& $\mathbf{68.6}$ & $\mathbf{77.9}$ & $\mathbf{79.3}$ \\

SAE & DINO
& $\underline{41.6}$ & $\underline{67.3}$ & $74.9$
& $60.0$ & $74.1$ & $76.8$ \\

SAE & --
& $37.9$ & $66.6$ & $\underline{75.0}$
& $65.7$ & $75.9$ & $\underline{79.1}$ \\

DINO & SAE
& $35.4$ & $66.3$ & $\underline{75.0}$
& $67.9$ & $\underline{77.4}$ & $78.8$ \\

DINO & DINO
& $34.9$ & $66.5$ & $74.9$
& $\underline{68.0}$ & $74.4$ & $75.4$ \\

DINO & --
& $31.8$ & $66.0$ & $74.8$
& $65.7$ & $74.1$ & $79.0$ \\

-- & DINO
& $38.1$ & $62.2$ & $71.3$
& $38.0$ & $61.8$ & $69.9$ \\

-- & SAE
& $37.5$ & $62.2$ & $71.6$
& $48.3$ & $62.8$ & $71.5$ \\

\bottomrule
\end{tabular}
}
\end{minipage}
\hfill
\begin{minipage}[t]{0.49\textwidth}
\centering
\captionof{table}{\textbf{Ablation study on the neighborhood size $k$.}
\textbf{Bold} and \underline{underlined} entries indicate the best and second-best results, respectively.}
\label{tab:k_ablation}

\vspace{3pt}
\centering
\small
\resizebox{\linewidth}{!}{
\begin{tabular}{@{}c ccc ccc ccc@{}}
\toprule
\multirow{2}{*}{$\boldsymbol{k}$}
& \multicolumn{3}{c}{\textbf{NABirds Acc}}
& \multicolumn{3}{c}{\textbf{Waterbirds WGA}}
& \multicolumn{3}{c}{\textbf{Waterbirds Acc}} \\
\cmidrule(lr){2-4}
\cmidrule(lr){5-7}
\cmidrule(lr){8-10}
& 10\% & 30\% & 50\%
& 10\% & 30\% & 50\%
& 10\% & 30\% & 50\% \\
\midrule

$1$
& $42.0$ & $\underline{67.4}$ & $74.7$
& $62.9$ & $\underline{75.6}$ & $\underline{78.8}$
& $91.9$ & $93.7$ & $94.3$ \\

$2$
& $42.4$ & $\mathbf{67.5}$ & $74.7$
& $63.2$ & $74.6$ & $76.8$
& $92.1$ & $93.8$ & $\mathbf{94.9}$ \\

$3$
& $\mathbf{42.9}$ & $\underline{67.4}$ & $\mathbf{75.2}$
& $68.6$ & $\mathbf{77.9}$ & $\mathbf{79.3}$
& $\underline{92.3}$ & $\underline{94.0}$ & $94.4$ \\

$5$
& $\underline{42.6}$ & $67.2$ & $\underline{74.9}$
& $\underline{69.0}$ & $71.8$ & $78.0$
& $92.1$ & $\underline{94.0}$ & $\underline{94.7}$ \\

$10$
& $41.1$ & $66.2$ & $74.5$
& $\mathbf{71.5}$ & $75.1$ & $76.3$
& $\mathbf{92.7}$ & $\mathbf{94.2}$ & $94.5$ \\

$20$
& $40.1$ & $66.5$ & $74.7$
& $62.8$ & $72.1$ & $76.6$
& $\mathbf{92.7}$ & $93.9$ & $94.2$ \\

\bottomrule
\end{tabular}
}

\captionof{table}{\textbf{Effect of normalization} over neighbor similarity.
We report accuracy for NABirds and worst-group accuracy (WGA)
for Waterbirds. The better result in each column is bolded.}
\label{tab:row_normalization_full}

\vspace{3pt}

\begingroup
\small
\setlength{\tabcolsep}{4pt}
\resizebox{0.95\textwidth}{!}{
\begin{tabular}{@{}l ccc ccc@{}}
\toprule
\multirow{2}{*}{\textbf{Assignment}}
& \multicolumn{3}{c}{\textbf{NABirds Acc}}
& \multicolumn{3}{c}{\textbf{Waterbirds WGA}} \\
\cmidrule(lr){2-4}
\cmidrule(lr){5-7}
& 10\% & 30\% & 50\%
& 10\% & 30\% & 50\% \\
\midrule

Unnormalized
& $42.6$ & $67.1$ & $74.9$
& $\mathbf{69.1}$ & $75.9$ & $78.7$ \\

Row-normalized
& $\mathbf{42.9}$ & $\mathbf{67.4}$ & $\mathbf{75.2}$
& $68.6$ & $\mathbf{77.9}$ & $\mathbf{79.3}$ \\

\bottomrule
\end{tabular}
}
\endgroup
\end{minipage}
\end{table}

\subsection{Concept and Local Representations}
Table~\ref{tab:feature_space_ablation} reports the per-rate results
underlying Figure~\ref{fig:feature_space_ablation}. Using SAE features for
both components achieves the best result. Removing the
concept component causes the largest drop, especially on Waterbirds at the
$10\%$ rate, where WGA falls from $68.6\%$ to $38.0\%$ and $48.3\%$.
The advantage of SAE-based concept coverage is most pronounced at low
selection rates: at $10\%$ on NABirds, it outperforms all DINO-based
concept variants by at least $6.2\%$.

\mypara{Effect of similarity normalization}
Table~\ref{tab:row_normalization_full} reports the per-rate results
underlying Table~\ref{tab:row_normalization}. Row normalization improves
performance in five of the six settings.

\subsection{Local Support Weight \texorpdfstring{$\beta$}{beta} and Neighborhood Size \texorpdfstring{$k$}{k}}

\begin{table}[h]
\caption{
\textbf{Sensitivity to the local support weight $\beta$.}
Comparison of performance across different local support weights and selection rates.
We report classification accuracy (Acc.) for all classification datasets and worst-group accuracy (WGA) for Waterbirds and Spawrious.
\textbf{Bold} and \underline{underlined} entries indicate the best and second-best results, respectively.
}
\label{tab:beta_ablation}
\vspace{-0.15cm}
\small
\setlength{\tabcolsep}{2.5pt}
\centering
\resizebox{\linewidth}{!}{
\begin{tabular}{c|ccc|ccc|ccc|ccc|ccc|ccc|ccc}
    \toprule[1pt]

    $\boldsymbol{\beta}$
    & \multicolumn{3}{c|}{\textbf{NABird Acc}}
    & \multicolumn{3}{c|}{\textbf{Stanford Cars Acc}}
    & \multicolumn{3}{c|}{\textbf{Waterbirds Acc}}
    & \multicolumn{3}{c|}{\textbf{Waterbirds WGA}}
    & \multicolumn{3}{c|}{\textbf{Spawrious Acc}}
    & \multicolumn{3}{c|}{\textbf{Spawrious WGA}}
    & \multicolumn{3}{c}{\textbf{Tiny-ImageNet Acc}}
    \\

    \cmidrule(lr){2-4}
    \cmidrule(lr){5-7}
    \cmidrule(lr){8-10}
    \cmidrule(lr){11-13}
    \cmidrule(lr){14-16}
    \cmidrule(lr){17-19}
    \cmidrule(lr){20-22}

    \textbf{Selection Rate}
    & \textbf{10\%} & \textbf{30\%} & \textbf{50\%}
    & \textbf{10\%} & \textbf{30\%} & \textbf{50\%}
    & \textbf{10\%} & \textbf{30\%} & \textbf{50\%}
    & \textbf{10\%} & \textbf{30\%} & \textbf{50\%}
    & \textbf{10\%} & \textbf{30\%} & \textbf{50\%}
    & \textbf{10\%} & \textbf{30\%} & \textbf{50\%}
    & \textbf{10\%} & \textbf{30\%} & \textbf{50\%}
    \\

    \midrule[1pt]

    $0$
    & $37.9$ & $66.6$ & $\underline{75.0}$
    & $17.8$ & $62.2$ & $78.6$
    & $91.9$ & $\mathbf{94.6}$ & $\mathbf{94.5}$
    & $65.7$ & $75.9$ & $79.1$
    & $\underline{87.8}$ & $\mathbf{87.3}$ & $\underline{87.1}$
    & $\mathbf{76.1}$ & $72.8$ & $71.9$
    & $12.0$ & $32.5$ & $45.1$
    \\

    $0.05$
    & $\mathbf{42.9}$ & $\mathbf{67.4}$ & $\mathbf{75.2}$
    & $23.9$ & $65.3$ & $\mathbf{78.8}$
    & $92.3$ & $94.0$ & $\underline{94.4}$
    & $\underline{68.6}$ & $\mathbf{77.9}$ & $\mathbf{79.3}$
    & $87.5$ & $87.1$ & $86.7$
    & $74.3$ & $74.7$ & $72.3$
    & $15.4$ & $37.0$ & $46.5$
    \\

    $0.1$
    & $\underline{42.8}$ & $67.2$ & $\underline{75.0}$
    & $24.2$ & $65.1$ & $\underline{78.7}$
    & $92.4$ & $93.9$ & $93.8$
    & $66.6$ & $71.7$ & $78.4$
    & $\mathbf{88.1}$ & $\underline{87.2}$ & $\mathbf{87.4}$
    & $74.8$ & $\mathbf{76.9}$ & $\underline{74.6}$
    & $15.7$ & $36.9$ & $46.6$
    \\

    $0.25$
    & $42.7$ & $\underline{67.3}$ & $74.5$
    & $24.4$ & $\mathbf{65.8}$ & $\underline{78.7}$
    & $\underline{92.5}$ & $94.2$ & $94.2$
    & $65.7$ & $\underline{76.5}$ & $\mathbf{79.3}$
    & $87.5$ & $87.0$ & $86.2$
    & $75.6$ & $75.9$ & $\mathbf{74.9}$
    & $16.8$ & $34.7$ & $45.1$
    \\

    $0.5$
    & $42.6$ & $67.0$ & $74.1$
    & $24.2$ & $\underline{65.6}$ & $78.2$
    & $\mathbf{92.8}$ & $\underline{94.3}$ & $94.3$
    & $\mathbf{69.8}$ & $75.4$ & $75.9$
    & $87.0$ & $87.1$ & $86.4$
    & $75.9$ & $75.1$ & $71.9$
    & $21.1$ & $38.3$ & $45.9$
    \\

    $1.0$
    & $42.3$ & $66.1$ & $73.7$
    & $\mathbf{25.3}$ & $65.2$ & $78.5$
    & $92.4$ & $94.2$ & $\underline{94.4}$
    & $65.4$ & $75.4$ & $74.8$
    & $87.1$ & $86.9$ & $86.5$
    & $\underline{76.0}$ & $\underline{76.3}$ & $74.3$
    & $\underline{22.3}$ & $\underline{39.2}$ & $\underline{48.7}$
    \\

    $2.0$
    & $41.9$ & $66.0$ & $73.3$
    & $\underline{25.1}$ & $65.4$ & $77.8$
    & $91.6$ & $93.7$ & $94.3$
    & $65.4$ & $67.6$ & $77.3$
    & $86.5$ & $86.3$ & $86.8$
    & $\underline{76.0}$ & $72.4$ & $73.4$
    & $\mathbf{23.5}$ & $\mathbf{40.4}$ & $\mathbf{50.0}$
    \\

    \bottomrule[1pt]
\end{tabular}
}
\end{table}
\mypara{Sensitivity to the local support weight $\beta$}\label{beta_abl_full}
The local support weight $\beta$ controls the balance between concept coverage
and local support.
Table~\ref{tab:beta_ablation} reports the full sweep underlying the summary in
Figure~\ref{fig:beta_sweep}.
The trend is consistent with the main paper's results: fine-grained and
spurious-correlation settings favor a moderate amount of local support, with
the strongest overall performance around $\beta=0.05$, whereas general image
classification benefits from a substantially larger weight, with the best
performance at $\beta=2$.
We use the corresponding dataset-group setting throughout the main experiments.

\mypara{Effect of neighborhood size $k$}
We vary the number of nearest neighbors over
$k\in\{1,2,3,5,10,20\}$ in Table~\ref{tab:k_ablation}.
Overall, $k=3$ provides the most consistent performance across NABirds and
Waterbirds, while larger neighborhoods do not yield systematic improvements.
We therefore use $k=3$ as the default setting throughout our experiments.

\subsection{Choice of Utility Function}
\begin{table}[h]
\caption{\textbf{Ablation study on the utility function.}
We compare different utility functions used in our selection objective.
Within each $\beta$ group, the best and second-best results in each
column are bolded and underlined, respectively.}
\label{tab:activation_ablation}

\centering
\small
\setlength{\tabcolsep}{7.0pt}
\resizebox{0.8\linewidth}{!}{
\begin{tabular}{@{}l l ccc ccc ccc@{}}
\toprule
\multirow{2}{*}{$\boldsymbol{\beta}$}
& \multicolumn{1}{c}{\multirow{2}{*}{\textbf{Utility}}}
& \multicolumn{3}{c}{\textbf{NABirds Acc}}
& \multicolumn{3}{c}{\textbf{Waterbirds Acc}}
& \multicolumn{3}{c}{\textbf{Waterbirds WGA}} \\
\cmidrule(lr){3-5}
\cmidrule(lr){6-8}
\cmidrule(lr){9-11}
& & 10\% & 30\% & 50\%
& 10\% & 30\% & 50\%
& 10\% & 30\% & 50\% \\
\midrule

\multirow{4}{*}{$0$}
& Linear
& $37.1$ & $61.6$ & $70.9$
& $89.7$ & $91.1$ & $92.7$
& $56.7$ & $61.1$ & $61.2$ \\

& Power
& $\mathbf{39.9}$ & $\underline{66.7}$ & $\underline{74.1}$
& $\underline{92.2}$ & $93.7$ & $\mathbf{94.6}$
& $\mathbf{68.3}$ & $\underline{74.4}$ & $\underline{77.3}$ \\

& Log
& $\underline{37.9}$ & $66.6$ & $\mathbf{75.0}$
& $91.9$ & $\mathbf{94.6}$ & $\underline{94.5}$
& $65.7$ & $\mathbf{75.9}$ & $\mathbf{79.1}$ \\

& Bounded exp.
& $\underline{37.9}$ & $\mathbf{66.8}$ & $\mathbf{75.0}$
& $\mathbf{93.4}$ & $\underline{94.0}$ & $\underline{94.5}$
& $\underline{67.1}$ & $74.1$ & $\mathbf{79.1}$ \\

\midrule

\multirow{4}{*}{$0.05$}
& Linear
& $37.8$ & $61.1$ & $69.9$
& $90.4$ & $90.9$ & $91.8$
& $50.6$ & $59.4$ & $61.2$ \\

& Power
& $\underline{40.3}$ & $\underline{66.6}$ & $74.1$
& $92.1$ & $93.0$ & $\underline{93.7}$
& $65.4$ & $65.3$ & $74.8$ \\

& Log
& $\mathbf{42.9}$ & $\mathbf{67.4}$ & $\mathbf{75.2}$
& $\underline{92.3}$ & $\mathbf{94.0}$ & $\mathbf{94.4}$
& $\underline{68.6}$ & $\mathbf{77.9}$ & $\mathbf{79.3}$ \\

& Bounded exp.
& $42.4$ & $66.1$ & $\underline{74.4}$
& $\mathbf{92.6}$ & $\underline{93.7}$ & $\mathbf{94.4}$
& $\mathbf{69.6}$ & $\underline{73.2}$ & $\underline{79.0}$ \\

\bottomrule
\end{tabular}
}
\end{table}

Our objective uses a logarithmic utility to impose diminishing returns on
accumulated concept support. We ablate this choice by replacing the logarithm
with a utility $\phi$:

\[
F_{\phi}(\mathcal{S})
=
\sum_{c=1}^{C}
\phi\!\left(
    \sum_{\mI_i\in \mathcal{S}}\tilde{\vx}_i[c]
\right).
\]

where $\tilde{x}_i[c]$ is the effective concept contribution defined in the Eq.~\ref{eq:local}. We compare four utilities:

\[
\begin{aligned}
\phi_{\mathrm{linear}}(t) &= t,
&\qquad
\phi_{\mathrm{power}}(t) &= 2\bigl(\sqrt{1+t}-1\bigr), \\
\phi_{\mathrm{log}}(t) &= \log(1+t),
&\qquad
\phi_{\mathrm{exp}}(t) &= 1-\exp(-t).
\end{aligned}
\]

All four satisfy $\phi(0)=0$ and $\phi'(0)=1$, giving the same initial
marginal value. The linear utility assigns constant value to additional
support, whereas the remaining utilities introduce diminishing returns;
the exponential utility additionally saturates at one.
Table~\ref{tab:activation_ablation} compares these choices with
$\beta=0$ and $\beta=0.05$.
Nonlinear utilities generally outperform the linear alternative, supporting
diminishing returns on accumulated concept support. We use the logarithmic
utility for its strong overall performance across datasets and selection rates.

\subsection{SAE Features vs.\ Selection Objective}
\begin{table}[h]
\caption{
\textbf{Geometric selection in DINO and SAE feature spaces.}
We compare K-Center using dense DINO features and sparse SAE features.The better result is \textbf{bolded}.
}
\label{tab:kcenter_sae}

\vspace{3pt}
\centering
\small

\resizebox{0.68\linewidth}{!}{
\begin{tabular}{@{}l cc cc@{}}
\toprule
\multirow{2}{*}{\textbf{Selection Rate}}
& \multicolumn{2}{c}{\textbf{NABirds Acc}}
& \multicolumn{2}{c}{\textbf{Waterbirds WGA}} \\
\cmidrule(lr){2-3}
\cmidrule(lr){4-5}
& K-Center (DINO) & K-Center (SAE)
& K-Center (DINO) & K-Center (SAE) \\
\midrule

10\%
& $31.6$ & $\mathbf{32.7}$
& $57.5$ & $\mathbf{61.2}$ \\

30\%
& $64.9$ & $\mathbf{66.5}$
& $64.2$ & $\mathbf{70.3}$ \\

50\%
& $74.1$ & $\mathbf{75.0}$
& $\mathbf{79.4}$ & $77.0$ \\

\bottomrule
\end{tabular}
}
\end{table}

Table 16 tests whether the gains come from the SAE representation or from our objective. Running K-Center on SAE features instead of DINO features improves it in five of six settings, yet K-Center (SAE) stays below our method in all six, and on NABirds at $10$\% it remains below Random ($32.7$ vs. $35.6$). At that setting, switching to SAE features adds $1.1$ points to K-Center, and our objective adds another $10.2$. The representation helps, and most of the gain comes from the objective.

\subsection{SAE Pretraining Data}
\begin{table}[h]
\caption{\textbf{Robustness to SAE pretraining data.}
We compare SAEs trained on the target dataset with an SAE trained on ImageNet-1K.
\textbf{Bold} and \underline{underlined} entries indicate the best and second-best results, respectively.}
\label{tab:sae_pretraining}

\vspace{3pt}
\centering
\small
\resizebox{0.8\linewidth}{!}{
\begin{tabular}{@{}l cc cc cc@{}}
\toprule
\multirow{2}{*}{\textbf{Selection Rate}}
& \multicolumn{2}{c}{\textbf{NABirds Acc}}
& \multicolumn{2}{c}{\textbf{Waterbirds WGA}}
& \multicolumn{2}{c}{\textbf{Waterbirds Acc}} \\
\cmidrule(lr){2-3}
\cmidrule(lr){4-5}
\cmidrule(lr){6-7}
& Target SAE & IN1K SAE
& Target SAE & IN1K SAE
& Target SAE & IN1K SAE \\
\midrule

10\%
& $\mathbf{43.0}$ & $\underline{42.9}$
& $\underline{63.4}$ & $\mathbf{68.6}$
& $\underline{91.7}$ & $\mathbf{92.3}$ \\

30\%
& $\mathbf{67.4}$ & $\mathbf{67.4}$
& $\underline{73.2}$ & $\mathbf{77.9}$
& $\underline{93.8}$ & $\mathbf{94.0}$ \\

50\%
& $\underline{74.4}$ & $\mathbf{75.2}$
& $\underline{78.7}$ & $\mathbf{79.3}$
& $\mathbf{95.1}$ & $\underline{94.4}$ \\

\bottomrule
\end{tabular}
}
\end{table}

Table~\ref{tab:sae_pretraining} compares SAEs trained on each target dataset
with a single SAE pretrained on ImageNet-1K.
Across NABirds and Waterbirds, the ImageNet-1K SAE performs comparably to target-trained SAEs.
This indicates that the learned sparse concept space transfers well across
datasets and does not require retraining an SAE for each new selection task.
We therefore use a single ImageNet-1K-pretrained SAE throughout our main
experiments.

\subsection{Downstream Architecture}
\begin{table*}[h]

\caption{
\textbf{Cross-architecture generalization.}
Comparison of subset selection performance using the same selected subsets with ResNet-50 and ViT-B/16.
We report classification accuracy (Acc.) for NABirds, and worst-group accuracy (WGA) and overall accuracy (Acc.) for Waterbirds; each cell shows ResNet-50 / ViT-B/16.
\textbf{Bold} and \underline{underlined} entries indicate the best and second-best results for each architecture, respectively.
}
\label{tab:cross_architecture}
\small
\setlength{\tabcolsep}{3pt}
\centering

\resizebox{\linewidth}{!}{
\begin{tabular}{l|ccc|ccc|ccc}
\toprule

\textbf{Method}
&\multicolumn{3}{c|}{\textbf{NABirds Acc}}
&\multicolumn{3}{c|}{\textbf{Waterbirds WGA}}
&\multicolumn{3}{c}{\textbf{Waterbirds Acc}}
\\

\cmidrule(lr){2-4}
\cmidrule(lr){5-7}
\cmidrule(lr){8-10}

\textbf{Selection Rate}
&\textbf{10\%}&\textbf{30\%}&\textbf{50\%}
&\textbf{10\%}&\textbf{30\%}&\textbf{50\%}
&\textbf{10\%}&\textbf{30\%}&\textbf{50\%}
\\

\midrule[1pt]

\rowcolor{gray!15}
Random
&$35.6$ / $40.2$
&$64.9$ / $65.9$
&$73.3$ / $73.8$
&$52.3$ / $54.3$
&$67.5$ / $\underline{66.0}$
&$68.8$ / $\underline{68.6}$
&$84.1$ / $76.7$
&$89.6$ / $85.8$
&$91.6$ / $85.9$
\\

K-Center
&$31.6$ / $33.7$
&$64.9$ / $65.0$
&$74.1$ / $73.5$
&$57.5$ / $21.7$
&$64.2$ / $49.5$
&$\mathbf{79.4}$ / $62.3$
&$89.4$ / $85.5$
&$89.9$ / $84.9$
&$91.7$ / $88.7$
\\

Forgetting
&$22.4$ / $26.9$
&$63.3$ / $63.6$
&$74.3$ / $73.7$
&$31.6$ / $3.9$
&$60.1$ / $25.6$
&$73.4$ / $54.4$
&$73.8$ / $75.7$
&$85.1$ / $81.7$
&$91.0$ / $89.8$
\\

EL2N
&$16.8$ / $21.0$
&$61.2$ / $61.0$
&$74.4$ / $73.4$
&$31.0$ / $30.4$
&$64.2$ / $38.2$
&$72.9$ / $63.2$
&$78.4$ / $79.9$
&$87.0$ / $85.0$
&$92.7$ / $89.7$
\\

D2 Pruning
&$25.8$ / $31.0$
&$\underline{66.6}$ / $66.6$
&$\underline{74.7}$ / $\underline{74.1}$
&$47.0$ / $50.5$
&$56.4$ / $37.5$
&$60.8$ / $47.2$
&$89.0$ / $\underline{88.6}$
&$\underline{92.2}$ / $87.9$
&$\underline{93.3}$ / $\underline{91.2}$
\\

Moderate
&$34.6$ / $37.7$
&$64.9$ / $66.1$
&$73.7$ / $73.8$
&$\underline{60.5}$ / $\underline{60.6}$
&$\underline{68.0}$ / $57.5$
&$71.1$ / $61.4$
&$\underline{91.2}$ / $88.0$
&$91.8$ / $\underline{90.7}$
&$92.5$ / $88.7$
\\

CCS
&$\underline{39.7}$ / $43.5$
&$65.5$ / $\underline{66.9}$
&$73.6$ / $73.6$
&$52.8$ / $38.8$
&$67.1$ / $55.9$
&$69.9$ / $64.5$
&$86.3$ / $73.6$
&$90.4$ / $86.4$
&$92.0$ / $87.0$
\\

InfoMax
&$22.5$ / $27.2$
&$64.7$ / $64.9$
&$72.7$ / $72.8$
&$22.6$ / $30.5$
&$61.7$ / $42.8$
&$70.9$ / $68.4$
&$74.4$ / $65.4$
&$88.3$ / $85.3$
&$91.8$ / $89.2$
\\
RAM-APL
&$39.3$ / $\underline{43.6}$
&$61.7$ / $64.5$
&$69.8$ / $71.6$
&$30.2$ / $13.9$
&$51.2$ / $43.0$
&$65.0$ / $24.1$
&$75.6$ / $73.7$
&$82.7$ / $80.7$
&$88.8$ / $85.3$
\\

\midrule

Ours 
&$\mathbf{42.9}$ / $\mathbf{44.9}$
&$\mathbf{67.4}$ / $\mathbf{67.6}$
&$\mathbf{75.2}$ / $\mathbf{74.4}$
&$\mathbf{68.6}$ / $\mathbf{63.7}$
&$\mathbf{77.9}$ / $\mathbf{68.2}$
&$\underline{79.3}$ / $\mathbf{80.4}$
&$\mathbf{92.3}$ / $\mathbf{92.2}$
&$\mathbf{94.0}$ / $\mathbf{91.9}$
&$\mathbf{94.4}$ / $\mathbf{91.8}$
\\

\bottomrule
\end{tabular}
}

\end{table*}

We evaluate the same selected subsets with both ResNet-50 and ViT-B/16.
As shown in Table~\ref{tab:cross_architecture}, the advantage of our subsets
is consistently preserved after switching the downstream architecture.
With ViT-B/16, our method achieves the best NABirds accuracy, Waterbirds WGA,
and Waterbirds overall accuracy across all three selection rates. Several baselines degrade much more than ours when the architecture changes. At $10$\%, Waterbirds WGA falls from $57.5$ to $21.7$ for K-Center and from $31.6$ to $3.9$ for Forgetting, while ours goes from 68.6 to 63.7.

\newpage

\section{Additional Results}\label{sec:additional_results}

\subsection{ImageNet-1K and Distribution Shift}
\begin{table}[h]
\caption{\textbf{ImageNet-1K classification and robustness results.}
We report top-1 accuracy (\%) on four robustness benchmarks and
the standard ImageNet-1K validation set at different selection rates.
The best and second-best results are bolded and underlined,
respectively. }
\label{tab:imagenet_robustness}

\vspace{3pt}
\centering
\small
\setlength{\tabcolsep}{3.5pt}

\resizebox{0.8\linewidth}{!}{
\begin{tabular}{@{}l ccc ccc ccc ccc ccc@{}}
\toprule
\multirow{2}{*}{\textbf{Method}}
& \multicolumn{3}{c}{\textbf{ImageNet-A}}
& \multicolumn{3}{c}{\textbf{ImageNet-R}}
& \multicolumn{3}{c}{\textbf{ImageNet-D}}
& \multicolumn{3}{c}{\textbf{ImageNet-V2}}
& \multicolumn{3}{c}{\textbf{ImageNet-1K Val}} \\

\cmidrule(lr){2-4}
\cmidrule(lr){5-7}
\cmidrule(lr){8-10}
\cmidrule(lr){11-13}
\cmidrule(lr){14-16}

& $10\%$ & $30\%$ & $50\%$
& $10\%$ & $30\%$ & $50\%$
& $10\%$ & $30\%$ & $50\%$
& $10\%$ & $30\%$ & $50\%$
& $10\%$ & $30\%$ & $50\%$ \\

\midrule

Random
& $\underline{1.4}$ & $1.2$ & $1.1$
& $15.8$ & $24.1$ & $\mathbf{25.9}$
& $\underline{2.9}$ & $\underline{6.3}$ & $6.8$
& $36.3$ & $\underline{49.3}$ & $\underline{50.9}$
& $48.6$ & $64.2$ & $67.2$ \\

EL2N
& $1.2$ & $1.0$ & $1.1$
& $9.3$ & $19.4$ & $23.2$
& $1.4$ & $3.6$ & $6.1$
& $13.4$ & $23.4$ & $45.2$
& $11.2$ & $45.4$ & $64.1$ \\

CCS
& $1.3$ & $\underline{1.6}$ & $\underline{1.4}$
& $\underline{16.0}$ & $\underline{24.5}$ & $\underline{25.1}$
& $2.7$ & $6.2$ & $\underline{6.9}$
& $\mathbf{37.4}$ & $49.1$ & $\mathbf{51.2}$
& $\mathbf{50.1}$ & $\underline{64.8}$ & $\mathbf{68.4}$ \\

Ours
& $\mathbf{1.7}$ & $\mathbf{2.0}$ & $\mathbf{1.5}$
& $\mathbf{16.4}$ & $\mathbf{25.8}$ & $\mathbf{25.9}$
& $\mathbf{3.3}$ & $\mathbf{6.8}$ & $\mathbf{7.6}$
& $\underline{37.2}$ & $\mathbf{50.1}$ & $50.7$
& $\underline{49.2}$ & $\mathbf{65.2}$ & $\underline{67.7}$ \\

\bottomrule
\end{tabular}
}
\end{table}

\mypara{Standard classification}
We evaluate ImageNet-1K subset selection at the $10\%$, $30\%$, and $50\%$
selection rates, comparing our method with Random, EL2N, and CCS under
identical downstream training settings.
As shown in Table~\ref{tab:imagenet_robustness}, our method improves
validation accuracy over random selection at all three rates, although the
gains are modest compared with those on the fine-grained,
spurious-correlation, and multimodal pretraining benchmarks. CCS is
competitive on this benchmark.

The smaller gains on ImageNet-1K may reflect differences in which visual
evidence limits learning across dataset regimes.
In fine-grained and spurious-correlation settings, under-supported concepts
and contextual variation are closely associated with where larger gains occur, as we discussed in \S\ref{sec:analysis}.
One possible explanation is that random ImageNet-1K subsets already preserve
much of the evidence needed for standard classification, leaving less room for
improvement through concept-support redistribution.
Exploring how concept support should be characterized and reallocated in broad,
general-purpose datasets is an important direction for future work.

\mypara{Robustness under distribution shift}
While our method does not establish a clear advantage on the standard validation set, we further compare the performance on robustness test sets. Specifically, we evaluate the same models on
ImageNet-A~\citep{hendrycks2021nae},
ImageNet-R~\citep{hendrycks2021many},
ImageNet-D~\citep{zhang2024imagenet_d}, and
ImageNet-V2~\citep{recht2019imagenet}.
Our method achieves the best accuracy on ImageNet-A, ImageNet-R, and
ImageNet-D at all three selection rates, while CCS performs better on
ImageNet-V2 at the $10\%$ and $50\%$ rates. The benefit of our selection is
therefore more pronounced under larger distribution shifts than on the
standard validation set, benefiting from broader concept coverage.

\subsection{Concept-Support Profiles of All Selectors}\label{sec:concept_profile}
Table~\ref{tab:concept-profile-all} extends the concept-support profile of Table~\ref{tab:concept-profile} to all baselines on $10\%$ Stanford Cars and NABirds subsets.
The trend is consistent across selectors and datasets: every baseline, whether geometry-based, score-based, or distribution-based, remains close to the random profile, whereas our selection substantially spreads support more evenly and shifts it toward less prevalent concepts.
\begin{table}[t]
    \caption{
    \textbf{Concept-support profiles of 10\% subsets for all baselines on (a) NABirds and (b) Stanford Cars.}
    Every baseline stays close to the random profile on all three measures, whereas our selection covers more concepts, spreads its support more evenly, and shifts support toward less prevalent concepts.
    }
    \label{tab:concept-profile-all-both}

    \centering
    \small
    \setlength{\tabcolsep}{3.0pt}
    \captionsetup[subtable]{position=bottom}
    \vspace{-4pt}

    \begin{subtable}[t]{0.49\linewidth}
        \centering
        \begin{tabular}{@{}l|ccc@{}}
            \toprule
            \textbf{Methods}
            & \shortstack{\textbf{Unique}\\\textbf{Coverage}}
            & \shortstack{\textbf{Effective}\\\textbf{Number}}
            & \shortstack{\textbf{Prevalence}\\\textbf{Ratio}} \\
            \midrule
            \textbf{Random}       & $7{,}973$ & $865$ & $1.00$ \\
            \midrule
            \textbf{K-Center}     & $8{,}215$ & $893$ & $0.97$ \\
            \textbf{Moderate}     & $8{,}112$ & $876$ & $0.99$ \\
            \textbf{CCS}          & $7{,}855$ & $846$ & $1.02$ \\
            \textbf{Forgetting}   & $8{,}120$ & $888$ & $0.97$ \\
            \textbf{EL2N}         & $8{,}181$ & $910$ & $0.94$ \\
            \textbf{D$^2$-Pruning}& $8{,}099$ & $887$ & $0.97$ \\
            \textbf{InfoMax}      & $8{,}205$ & $902$ & $0.95$ \\
            \textbf{RAM-APL}      & $7{,}449$ & $790$ & $1.09$ \\
            \textbf{SDM}          & $7{,}918$ & $891$ & $0.95$ \\
            \midrule
            \textbf{Ours}         & $\mathbf{8{,}951}$ & $\mathbf{1{,}124}$ & $\mathbf{0.74}$ \\
            \bottomrule
        \end{tabular}
        \caption{NABirds}
        \label{tab:concept-profile-all}
    \end{subtable}
    \hfill
    \begin{subtable}[t]{0.49\linewidth}
        \centering
        \begin{tabular}{@{}l|ccc@{}}
            \toprule
            \textbf{Methods}
            & \shortstack{\textbf{Unique}\\\textbf{Coverage}}
            & \shortstack{\textbf{Effective}\\\textbf{Number}}
            & \shortstack{\textbf{Prevalence}\\\textbf{Ratio}} \\
            \midrule
            \textbf{Random}       & $9{,}321$ & $1{,}491$ & $1.00$ \\
            \midrule
            \textbf{K-Center}     & $9{,}320$ & $1{,}481$ & $1.00$ \\
            \textbf{Moderate}     & $9{,}261$ & $1{,}477$ & $1.01$ \\
            \textbf{CCS}          & $9{,}104$ & $1{,}468$ & $1.02$ \\
            \textbf{Forgetting}   & $9{,}219$ & $1{,}516$ & $0.99$ \\
            \textbf{EL2N}         & $9{,}296$ & $1{,}532$ & $0.97$ \\
            \textbf{D$^2$-Pruning}& $9{,}304$ & $1{,}480$ & $1.01$ \\
            \textbf{InfoMax}      & $9{,}271$ & $1{,}522$ & $0.98$ \\
            \textbf{RAM-APL}      & $8{,}792$ & $1{,}390$ & $1.06$ \\
            \textbf{SDM}          & $9{,}026$ & $1{,}544$ & $0.95$ \\
            \midrule
            \textbf{Ours}         & $\mathbf{9{,}510}$ & $\mathbf{1{,}853}$ & $\mathbf{0.77}$ \\
            \bottomrule
        \end{tabular}
        \caption{Stanford Cars}
        \label{tab:concept-profile-all-cars}
    \end{subtable}
\end{table}

\subsection{Additional SAE Concept Visualizations}

\begin{figure}[!ht]
    \centering
    \includegraphics[width=0.62\linewidth]{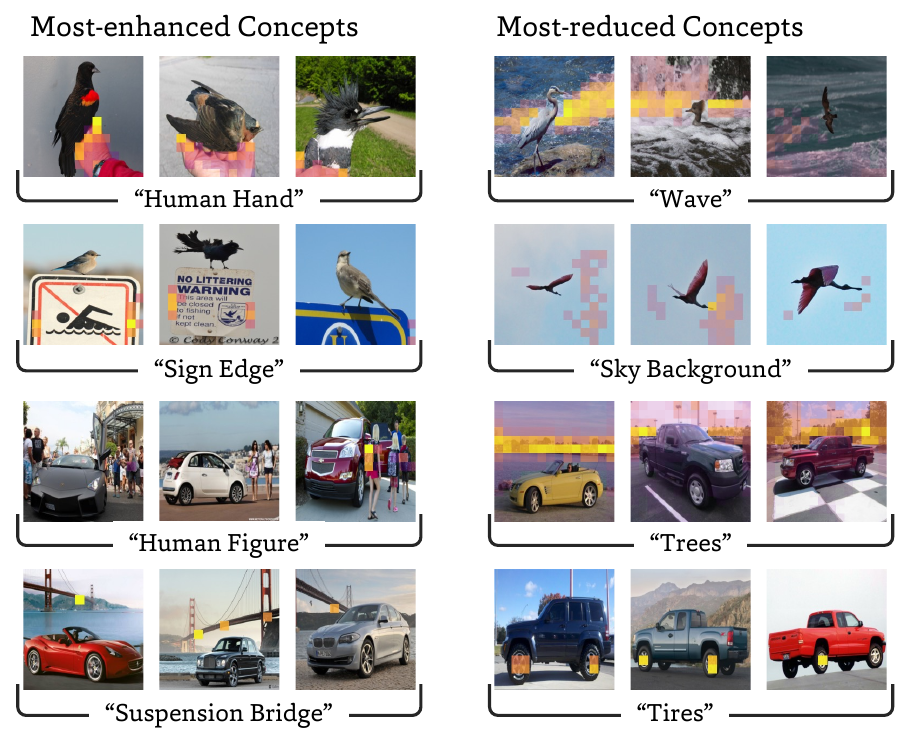}
    \vspace{-12pt}
    \caption{
    \textbf{Additional examples of enhanced and reduced SAE concepts.}
    We show representative images from concepts whose support changes most under
    our selection relative to random selection. Highlighted patches indicate high-activation regions.
    }
    \label{fig:sae_vis}
    \vspace{-12pt}
\end{figure}

Figure~\ref{fig:sae_vis} provides additional examples of concepts whose support
is most enhanced or reduced by our selection.
Enhanced concepts include uncommon contextual evidence such as human hands,
sign edges, human figures, and suspension bridges, while reduced concepts
include recurring visual patterns such as waves, sky backgrounds, trees, and
tires. These examples illustrate that our selection redistributes a fixed training
budget across visual factor. Support is shifted away from repeatedly represented patterns toward evidence
that is less frequently retained under random selection.

\newpage

\subsection{Additional Finer-CAM Visualizations}
\begin{figure}[!ht]
    \centering
    \includegraphics[width=\linewidth]{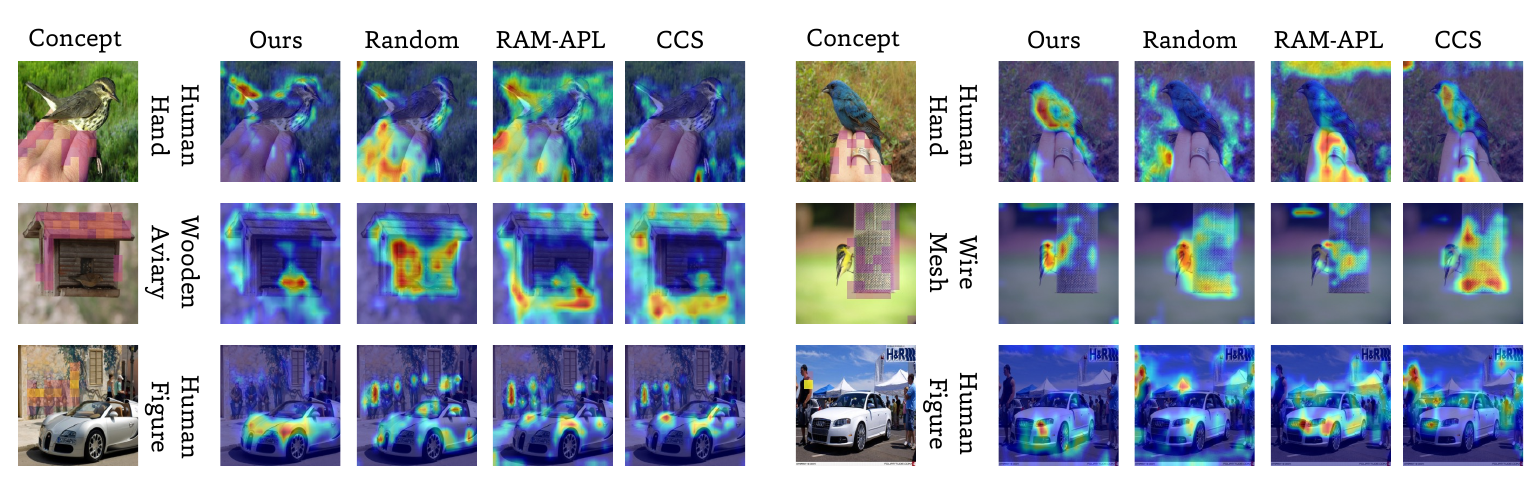}
    \vspace{-12pt}
    \caption{
    \textbf{Additional Finer-CAM visualizations on NABirds and Stanford Cars.} We show attribution maps for models trained on subsets
    selected by our method, Random, RAM-APL, and CCS. The \emph{Concept} column highlights image patches with high activation for the
    corresponding SAE latent.
    }
\label{fig:finer_cam_additional}
    \label{fig:finercam_vis}
    \vspace{-12pt}
\end{figure}

Figure~\ref{fig:finer_cam_additional} provides additional Finer-CAM examples
to complement the qualitative analysis in Figure~\ref{fig:concept_finer_cam}.
Across multiple enhanced concepts, models trained on our subsets
consistently place more attribution on the target object,
while models trained on random and prior works' subsets more often emphasize
the surrounding contextual evidence.

\newpage

\section{Relation to Concept-Based Data Selection}\label{sec:related_extended}

This work studies data selection from the perspective of how a limited training
budget is distributed across visual concepts. This direction is complementary
to recent advances that use concepts as signals for selecting individual
examples.~\cite{yang2025diversity} use SAE features to promote
feature diversity,~\cite{mehra2025model} estimate sample difficulty
through concept bottlenecks, and SAE-V~\citep{lou2025saev} uses concept-level
weights for multimodal filtering. These works demonstrate that concept
representations can provide useful semantic information for data selection.
Our work considers concept support at the level of the selected subset.
We characterize how limited training support is distributed across concepts
and how this allocation changes with subset selection.
Because each image contains multiple visual factors, selecting diverse samples
can still repeatedly reinforce already common evidence.
Our results show that making this allocation explicit helps reveal where the
training budget is spent and how these changes relate to downstream behavior.

A parallel line of work uses concept representations to align selection with a
desired target. MONA~\citep{ma2025task} selects examples according to a
target-task prototype, while SDM~\citep{das2026matched} matches a reference
concept distribution. These approaches are complementary to our setting. Related methods also differ in how the subset
is constructed: D$^2$-Pruning~\citep{maharana2024d2} propagates scalar difficulty
information over a sample graph, whereas~\cite{zhou2026dynamic}
control concept exposure dynamically by changing training samples across
epochs. We instead study concept support as a property of a fixed selected
subset. This perspective allows us to examine not only whether concept-aware
selection improves performance, but also how changing the support of particular
visual evidence is associated with fine-grained recognition, contextual
robustness, and multimodal pretraining.

\end{document}